\documentclass[lettersize,journal]{IEEEtran}
\usepackage{amsmath,amsfonts}
\usepackage{algorithmic}
\usepackage{algorithm}
\usepackage{array}
\usepackage[caption=false,font=normalsize,labelfont=sf,textfont=sf]{subfig}
\usepackage{textcomp}
\usepackage{stfloats}
\usepackage{url}
\usepackage{verbatim}
\usepackage{graphicx}
\usepackage{cite}
\usepackage{algorithm}
\usepackage{algorithmic}

\usepackage{colortbl}  
\usepackage{xcolor}
\usepackage{array}

\usepackage{graphicx}
\usepackage{multirow}
\usepackage{bbding}
\usepackage{booktabs} 
\usepackage{xcolor}
\newcommand{\eg}{\textit{e}.\textit{g}.}

\newcommand{\ie}{\textit{i}.\textit{e}.}

\definecolor{hollywoodcerise}{rgb}{0.96, 0.0, 0.63}
\definecolor{lasallegreen}{rgb}{0.03, 0.47, 0.19}
\definecolor{hanpurple}{rgb}{0.32, 0.09, 0.98}
\definecolor{green(pigment)}{rgb}{0.0, 0.65, 0.31}
\usepackage[pagebackref=true,breaklinks=true,letterpaper=true,colorlinks,bookmarks=false]{hyperref}
\hypersetup{colorlinks,linkcolor={red},citecolor={hanpurple},urlcolor={red}}  
\begin{document}

\title{360SFUDA++: Towards Source-free UDA for Panoramic Segmentation by Learning Reliable Category Prototypes}

\author{Xu Zheng,~\IEEEmembership{Student Member,~IEEE,}
        Pengyuan Zhou,
        Athanasios V. Vasilakos~\IEEEmembership{Senior Member,~IEEE,} \\
        Lin Wang*,~\IEEEmembership{Member,~IEEE,}\thanks{* Corresponding Author}
\thanks{Xu Zheng is with the AI Thrust, HKUST(GZ), Guangdong, China. E-mail: zhengxu128@gmail.com. Pengyuan Zhou is with the Aarhus University, Denmark. E-mail:pengyuan.zhou@ece.au.dk. Athanasios V. Vasilakos is with the University of Agder, Norway. E-mail: th.vasilakos@gmail.com.
Lin Wang is with AI/CMA Thrust, HKUST(GZ) and Dept. of CSE, HKUST, Hong Kong SAR, China, E-mail: linwang@ust.hk.
}
}

\markboth{Journal of \LaTeX\ Class Files,~Vol.~14, No.~8, August~2021}%
{Shell \MakeLowercase{\textit{et al.}}: A Sample Article Using IEEEtran.cls for IEEE Journals}


\maketitle

\begin{abstract}
In this paper, we address the challenging source-free unsupervised domain adaptation (SFUDA) for pinhole-to-panoramic semantic segmentation, 
given only a pinhole image pre-trained model (\ie, source) and unlabeled panoramic images (\ie, target). 
Tackling this problem is non-trivial due to three critical challenges: 1) semantic mismatches from the distinct Field-of-View (FoV) between domains, 2) style discrepancies inherent in the UDA problem, and 3) inevitable distortion of the panoramic images. To tackle these problems, we propose \textbf{360SFUDA++} that effectively extracts knowledge from the source pinhole model with only unlabeled panoramic images and transfers the reliable knowledge to the target panoramic domain.
Specifically, we first utilize Tangent Projection (TP) as it has less distortion and meanwhile slits the equirectangular projection (ERP) to patches with fixed FoV projection (FFP) to mimic the pinhole images. Both projections are shown effective in extracting knowledge from the source model. 
However, as the distinct projections make it less possible to directly transfer knowledge between domains, we then propose Reliable Panoramic Prototype Adaptation Module (RP$^2$AM) to transfer knowledge at both prediction and prototype levels. RP$^2$AM selects the confident knowledge and integrates panoramic prototypes for reliable knowledge adaptation. 
Moreover, we introduce Cross-projection Dual Attention Module (CDAM), which better aligns the spatial and channel characteristics across projections at the feature level between domains. 
Both knowledge extraction and transfer processes are synchronously updated to reach the best performance. Extensive experiments on the synthetic and real-world benchmarks, including outdoor and indoor scenarios, demonstrate that our 360SFUDA++ achieves significantly better performance than prior SFUDA methods. Project Page: \textcolor{red}{\url{https://vlislab22.github.io/360SFUDA/}}

\end{abstract}


\section{Introduction}
\label{Sec:intro}

\IEEEPARstart{T}{he} comprehensive scene perception capabilities afforded by $360^\circ$ cameras have rendered them exceedingly popular for a diverse array of applications, including autonomous driving and virtual reality~\cite{360survey}. Unlike pinhole cameras, which capture 2D planar images within a confined field-of-view (FoV), $360^\circ$ cameras provide an expansive FoV spanning $360^\circ \times 180^\circ$. Consequently, there has been a concerted research endeavors towards panoramic semantic segmentation~\cite{yang2019pass, yang2020omnisupervised, zhang2022bending, p2pda, PCS}, aiming at achieving comprehensive scene understanding for intelligent systems.

\begin{figure}[t!]
    \centering
    \includegraphics[width=\linewidth]{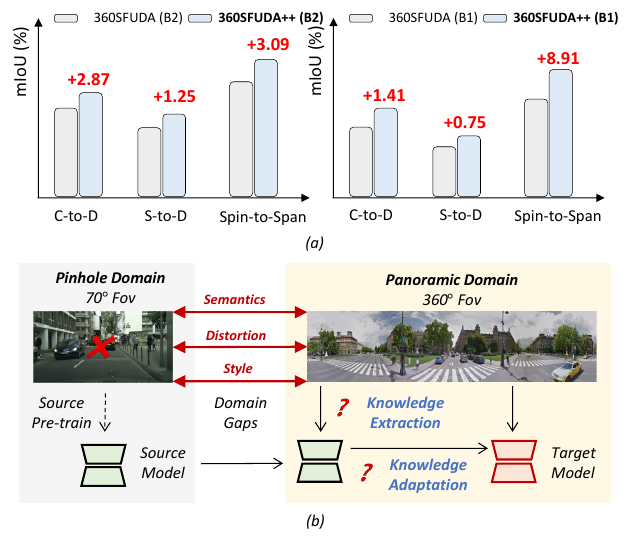}
    \vspace{-20pt}
    \caption{(a) Performance comparison between ~\cite{zheng2024semantics} and 360SFUDA++; (b) Prototype comparison between ~\cite{zheng2024semantics} and 360SFUDA++ on outdoor C-to-D scenario, different colors stand for different categories.
    }
    \label{fig:cover_fig}
\end{figure}

Generally, data captured by $360^\circ$ cameras in spherical form is commonly projected onto 2D planar representations, such as the Equirectangular Projection (ERP), to integrate seamlessly into existing imaging pipelines~\cite{360survey}, while still preserving omnidirectional information\footnote{Throughout this paper, the terms "omnidirectional" and "panoramic" are used interchangeably, with ERP images often denoting panoramic images.}.
However, ERP images always suffer from the inevitable distortion and object deformation due to the non-uniformly distributed pixels~\cite{zheng2023both}. Meanwhile, learning effective panoramic segmentation models is often hindered by the lack of large precisely labeled datasets due to the difficulty of annotation. 
For these reasons, some unsupervised domain adaptation (UDA) methods~\cite{zhang2022bending, zhang2022behind, zheng2023both} have been proposed to transfer the knowledge from the pinhole image domain to the panoramic image domain. In some crucial application scenarios, \eg, autonomous driving, source datasets are not always accessible due to privacy and commercial issues, such as data portability and transmission costs. A notable example is the recent large-scale model, SAM~\cite{kirillov2023segment}, which has made significant advancements in instance segmentation for pinhole images. However, the size of the source datasets (10TB) renders them impractical for reuse in downstream tasks~\cite{kirillov2023segment}.

In this paper, we probe an interesting yet challenging problem -- \textit{source-free UDA (SFUDA) for panoramic segmentation, in which only the source model (pretrained with pinhole images) and unlabeled panoramic images are available}. As shown in Fig.~\ref{fig:cover_fig} (b), different from existing SFUDA methods,~\eg, \cite{liu2021source,ye2021source,yang2022source} for the pinhole-to-pinhole image adaptation, transferring knowledge from the pinhole-to-panoramic image domain is hampered by:
\textbf{ 1)} semantic mismatches arising from the disparate FoV between pinhole and $360^\circ$ cameras, namely 70$^\circ$ vs. 360$^\circ$; 
\textbf{2)} inevitable inherent distortion of the ERP; 
\textbf{3)} style discrepancies caused by the distinct camera sensors and captured scenes. In Tab.~\ref{Tab:City2PASS}, we demonstrate that simply applying existing SFUDA methods to the specific pinhole-to-panoramic adaptation problem yields only marginal performance enhancements.

To this end, we propose 360SFUDA++ that effectively extracts knowledge from the source pinhole model with only unlabeled panoramic images and transfers the reliable knowledge to the target panoramic domain.
\textit{Our key idea is to leverage the multi-projection versatility of 360$^\circ$ data for efficient and reliable domain knowledge transfer}.
\textbf{360SFUDA++} enjoys two key technical contributions. 
Specifically, we leverage Tangent Projection (TP)\footnote{See Algorithm \textcolor{red}{2} in the supplementary material for tangent projection.} and divide the ERP images into patches with a fixed FoV, dubbed Fixed FoV Projection (FFP), to mimic pinhole images and extract knowledge from the source model, aiming at alleviating distortion and semantic mismatches between domains.
Both projection techniques enable efficient knowledge extraction from the source model.
Considering the distinct projections make it hardly possible to directly transfer the extracted knowledge to the target model, we introduce the reliable panoramic prototype adaptation module (RP$^2$AM) (Sec.~\ref{RPPAM}) to achieve knowledge transfer at the prototype level (See Fig.~\ref{fig:cover_fig}). 
RP$^2$AM employs cross-projection prediction assessment (Sec.~\ref{CPPP}) to distinguish confident and uncertain pixels across three projections, extracting prototypes in the target panoramic domain (Sec.~\ref{CPPE}). The global prototypes from the source model with TP and FFP images are updated iteratively during adaptation to enhance knowledge extraction. Additionally, RP$^2$AM fine-tunes the source model using prototypes from FFP images to improve knowledge extraction (Sec.~\ref{FSMBKE}). Aligning prototypes from each FFP image enhances the model's awareness of distortion and semantics across the FoV.

For efficient knowledge adaptation, we apply prediction-level and prototype-level loss constraints to facilitate knowledge transfer to the unlabeled target panoramic domain (Sec.~\ref{Sec:3.3}). FFP predictions are reconstructed as pseudo labels for the target model. Prototype-level loss constraints are enforced in both intra-projection and cross-projection prototypical adaptation. Intra-projection prototypical adaptation occurs between confident and uncertain prototypes obtained with ERP images. Cross-projection prototypical adaptation utilizes global panoramic prototypes extracted from TP and FFP images from the source model, continually updated throughout the adaptation process to supervise confident prototypes in the target domain.

Moreover, the knowledge derived from the source model goes beyond predictions and prototypes, encompassing high-level features that encapsulate crucial image characteristics, significantly enhancing the performance of the target model. To address this, we introduce the Cross-projection Dual Attention Module (CDAM) (Sec.~\ref{CDAM}), aiming at harmonizing spatial and channel characteristics across projections and domains, maximizing the utilization of knowledge from the source model and mitigating the style discrepancy problems. CDAM reconstructs source model features from FFP images to provide a panoramic understanding of the surrounding environment, aligning them with ERP features from the target model to facilitate effective knowledge transfer.

We conduct extensive experiments on both synthetic and real-world benchmarks, including outdoor and indoor scenarios.
We adapt the state-of-the-art (SoTA) SFUDA methods~\cite{liu2021source, zhang2021prototypical, kundu2021generalize, huang2021model, yang2022source, guo2022simt} -- designed for pinhole-to-pinhole image adaptation -- to our problem in addressing the panoramic semantic segmentation.
As shown in Fig.~\ref{fig:cover_fig}, the results show that our framework significantly outperforms these methods by large margins of +1.25\%, +2.87\%, and +3.09\% on three benchmarks. 
We also evaluate our method against UDA methods~\cite{zhang2022behind, zhang2022bending, zheng2023both, zheng2023look}, using source data, the results demonstrate its comparable performance.

In summary, our main contributions are as follows: (\textbf{I}) we propose 360SFUDA++, which incorporates RP$^2$AM and CDAM to address the challenging source-free UDA task from pinhole to panoramic images; (\textbf{II}) The RP$^2$AM is proposed to achieve the panoramic knowledge adaptation with reliable prototypes between domains; (\textbf{III}) CDAM aims to harmonize spatial and channel characteristics across different projections of sphere data, aiming at mitigating the style discrepancy problem. (\textbf{IV}) Extensive experiments are conducted on both synthetic and real-world benchmarks (including indoor and outdoor scenarios) demonstrate the superiority of our 360SFUDA++. 


This work is an improvement over our CVPR 2024 work~\cite{zheng2024semantics}, achieved by substantially extending the method and experiment in the following ways.
\textbf{(I)} We introduced the cross-projection pixel-wise prediction assessment in RP$^2$AM to discern confident and uncertain predictions for better knowledge adaptation (Sec.~\ref{CPPP}); 
\textbf{(II)} We upgraded the prototype extraction to cross-projection prototype extraction to obtain confident and uncertain prototypes for better knowledge transfer (Sec.~\ref{CPPE}); 
\textbf{(III)} We added the loss constraints between uncertain and confident prototypes to achieve intra-projection knowledge adaptation (Sec.~\ref{CPPE});
\textbf{(IV)} We aligned the confident prototypes with the global panoramic prototypes for accurate and reliable prototypical knowledge adaptation (Sec.~\ref{Sec:3.3}); 
\textbf{(V)} We conducted more comparison experiments between the new framework and the previous version, as well as the existing SoTA SFUDA methods (Sec.~\ref{Experimental Results}); 
\textbf{(VI)} We validated the effectiveness of the introduced strategies and components with extensive quantitative and qualitative analysis (Sec.~\ref{Ablation Study}).

\begin{figure*}[h!]
    \centering
    \includegraphics[width=\textwidth]{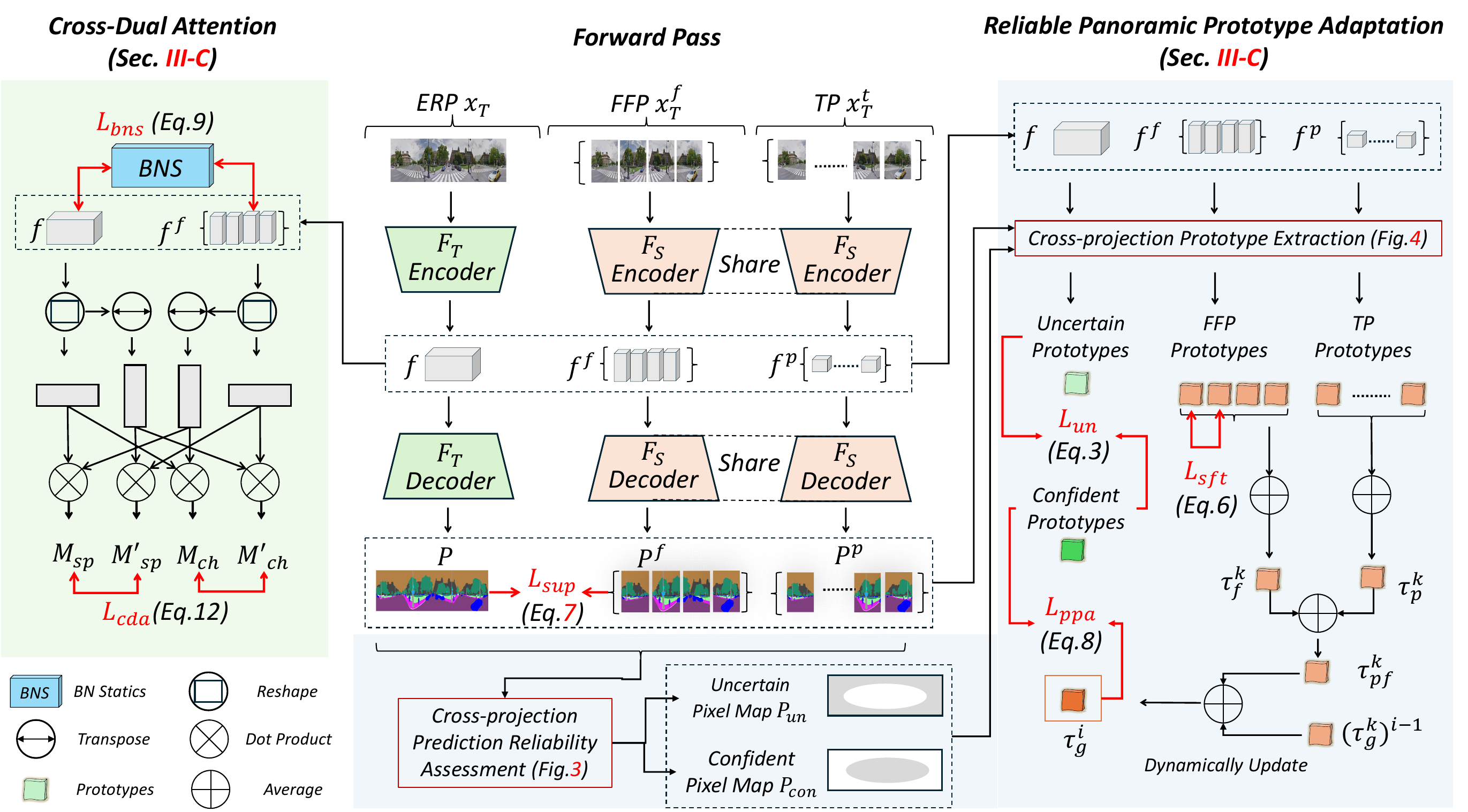}
    \vspace{-12pt}
    \caption{Overall framework of our proposed 360SFUDA++. 
    }
    \vspace{-12pt}
    \label{fig:overall}
\end{figure*}

\section{Related Work}
\subsection{Source-free UDA for Segmentation}
UDA aims to mitigate the impact of domain shift caused by data distribution discrepancies in downstream computer vision tasks, such as semantic segmentation~\cite{zhang2024goodsam,zhang2019category,chen2019domain,hoyer2022daformer,vu2019dada,zou2018unsupervised,stan2021unsupervised,fleuret2021uncertainty,shen2021unsupervised,vu2019advent,pan2020unsupervised,araslanov2021self, zheng2022uncertainty, chen2022uncertainty, zhu2023good, zheng2022transformer, chen4617170frozen, zheng2023distilling, chen2023clip, xie2023adversarial}. However, the source domain data may not always be accessible due to the privacy protection and data storage concerns. Intuitively, source-free UDA (SFUDA)~\cite{zheng2024eventdance,kundu2021generalize,yeh2021sofa,huang2021model} methods are proposed to adapt source models to a target domain without access to the source data. 
Existing source-free UDA methods can be categorized into two distinct approaches: those generating images independently of the source data, and those employing self-supervision techniques with target pseudo-labels.~\cite{ye2021source,yang2022source,kundu2021generalize,zhao2022source,liu2021source,bateson2022source,tian2024self,sun2024you}.
For instance, \cite{tian2024self} introduces a dual-branch collaborative learning framework for SFUDA which separate target data into confident samples and uncertain samples to figure out unreliable pixel-wise segmentation. 
In this paper, we attempt to achieve SFUDA from the pinhole image domain to the panoramic image domain. This task is nontrivial to be tackled due to the semantic mismatches, style discrepancies, and inevitable distortion of panoramic images~\cite{zheng2024semantics}.
Unlike these methods that focus on the source domain data estimation~\cite{liu2021source, ye2021source}, we propose 360SFUDA++, a novel SFUDA framework that effectively extracts knowledge from the source model with only panoramic images and transfers the knowledge to the target panoramic image domain. Different from the prior prototypical adaptation methods, such as ~\cite{tian2024self} which only focuses the confident and unreliable problem at sample level, our 360SFUDA++ subtly utilize the multi-projection versatility of panoramic data and achieve reliable prototypical adaptation in the pixel-level.
Experiments also show that naively applying these methods leads to less optimal performance (See Tab.~\ref{Tab:City2PASS}).

\subsection{UDA for Panoramic Semantic Segmentation}
It can be classified into three types, including adversarial training~\cite{Hoffman2018CyCADACA,Choi2019SelfEnsemblingWG,Sankaranarayanan2018LearningFS,Tsai2018LearningTA,zheng2023both}, pseudo labeling~\cite{Liu2021PanoSfMLearnerSM,Zhang2021DeepPanoContextP3,Wang2021DomainAS,Zhang2017CurriculumDA} and prototypical adaptation methods~\cite{zhang2022bending,zhang2022behind}. Specifically, the first line of research applies alignment approaches to capture the domain invariant characteristics of images~\cite{Hoffman2018CyCADACA,Li2019BidirectionalLF,Murez2018ImageTI}, feature~\cite{Hoffman2018CyCADACA,Chen2019ProgressiveFA,Hoffman2016FCNsIT,zheng2023both} and predictions~\cite{Luo2019TakingAC,MelasKyriazi2021PixMatchUD}. The second type of method generates pseudo labels for the target domain training. The last line of research, \eg, Mutual Prototype Adaption (MPA)~\cite{zhang2022bending}, utilizes the prototypical adaptation and mutually aligns the high-level features with the prototypes between domains. However, these methodologies often treat panoramic images as pinhole images during prototype extraction and adaptation, thereby neglecting the nuanced semantic details, object correspondences, and distortion inherent in panoramic FoV. In this paper, we address the SFUDA problem for panoramic segmentation by improving our previous work 360SFUDA~\cite{zheng2024semantics}.
Considering the distinct projection discrepancies between source and target domains, we propose an RP$^2$AM to transfer knowledge at the prototype level. Differently, we take the cross-projection pixel-wise predictions' confidence into consideration and further extract reliable prototypes for better knowledge adaptation.

\section{Methodology}   
\subsection{Overview}
The overall framework of our 360SFUDA++ is shown in Fig.~\ref{fig:overall}. 
Given only the source model $F_S$ and unlabeled panoramic image data $D_T$, the SFUDA objective is to learn a target model $F_T$ that effectively transfers knowledge from $F_S$ to the common $K$ categories across both domains.
Different from the traditional pinhole image-to-image adaptation~\cite{liu2021source,ye2021source,yang2022source}, the knowledge adaptation from pinhole to panoramic image domains is notably challenged by three primary factors, specifically:
1) semantic mismatch between pinhole and panoramic images caused by the FoV variations (70$^\circ$ vs. 360$^\circ$), which means there are more objects and cross-object correlations in panoramic images with 360$^\circ$ FoV; 
2) inevitable distortion and deformation in ERP of panoramic images; and 3) ubiquitous style discrepancies in almost all UDA tasks.

Therefore, naively applying existing SFUDA methods exhibits sub-optimal segmentation performance (See Tab.~\ref{Tab:City2PASS}), while UDA methods with source data, \eg, \cite{liu2021source} for panoramic segmentation fail to address the semantic discrepancies between the pinhole and panoramic images. Intuitively, the key challenges are
: \textbf{1)} how to extract knowledge from the source pinhole model with only unlabeled panoramic images and \textbf{2)} how to efficiently transfer knowledge to the target panoramic image domain. Overall, in our 360SFUDA++, \textbf{the key idea} \textit{is to leverage the multi-projection versatility of 360$^\circ$ data for efficient domain knowledge transfer.}

To tackle the first challenge (Sec.~\ref{Sec.3.2}), we utilize the Tangent Projection (TP) which is distinguished by its ability to mitigate distortion problems more effectively than ERP images~\cite{eder2020tangent}, leveraging the reduced distortion characteristics of TP to extract knowledge from the source pinhole model. Concurrently, we segment ERP images into discrete patches, each possessing a constant FoV to mimic the pinhole images, dubbed Fixed FoV Projection (FFP). Both TP and FFP facilitate the effective extraction of knowledge from the source model. However, the inherent differences in projection formats preclude a direct transfer of knowledge between domains. To overcome this obstacle, we introduce the Reliable Panoramic Prototype Adaptation Module (RP$^2$AM), designed to generate panoramic prototypes suitable for knowledge adaptation.
To address the second challenge (Sec.~\ref{Sec:3.3}), we first impose prediction and prototype level loss constraints, and propose a Cross-Dual Attention Module (CDAM) at the feature level to transfer knowledge and further address the style discrepancies between pinhole and panoramic images.

\subsection{Knowledge Extraction}
\label{Sec.3.2}
As illustrated in Fig.~\ref{fig:overall}, given the unlabled target domain (\ie, panoramic domain) ERP images $D_T=\{x_T|x_T \in \mathbf{R}^{H\times W\times 3}\}$, we first project them into TP images $D_T^{t}=\{x_T^{t}|x_T^{t} \in \mathbf{R}^{h\times w\times 3}\}$ and FFP images $D_T^{f}=\{x_T^{f}|x_T^{f} \in \mathbf{R}^{H\times W/4\times 3}\}$. Note that one ERP image corresponds to $18$ TP images as ~\cite{li2022omnifusion, zheng2023both} and $4$ FFP images with a fixed FoV of $90^\circ$ (ablation can be found in Sec.~\ref{Sec:ab}).
To obtain the high-level features and predictions from the source model for knowledge adaptation, TP and FFP images are first fed into the source model with batch sampling:
\begin{equation}
    P^p, f^p =F_S(x_T^{t}), \qquad P^f, f^f =F_S(x_T^{f}),
\end{equation}
where $f^p$, $f^f$, $P^p$, and $P^f$ are the source model features and predictions of the input TP and FFP images, respectively. 
For the target domain panoramic images, $x_T$ is fed into $F_T$ to obtain the target model features $f$ and predictions $P$ of the input batch of ERP images as $P, f = F_T(x_T)$.
To overcome inherent differences caused by the projection formats, we introduce the Reliable Panoramic Prototype Adaptation Module (RP$^2$AM) to obtain panoramic prototypes for effective and reliable knowledge adaptation.  


\subsubsection{Reliable Panoramic Prototype Adaptation Module}
\label{RPPAM}
In contrast to the previous version of this work~\cite{zheng2024semantics}, namely panoramic prototype adaptation module (PPAM), which posses three distinct characteristics, including: (a) class-wise prototypes are obtained from distinct projections directly to achieve knowledge transfer; (b) global prototypes are iteratively updated across training; and (c) hard pseudo labels are softened in the feature space to take full use of the source knowledge. 
Deviating from the previous work~\cite{zheng2024semantics}, we put forward a better prototypical adaptation strategy with more reliable knowledge extraction and transfer. 

\begin{figure}[t!]
    \centering
    \includegraphics[width=0.49\textwidth]{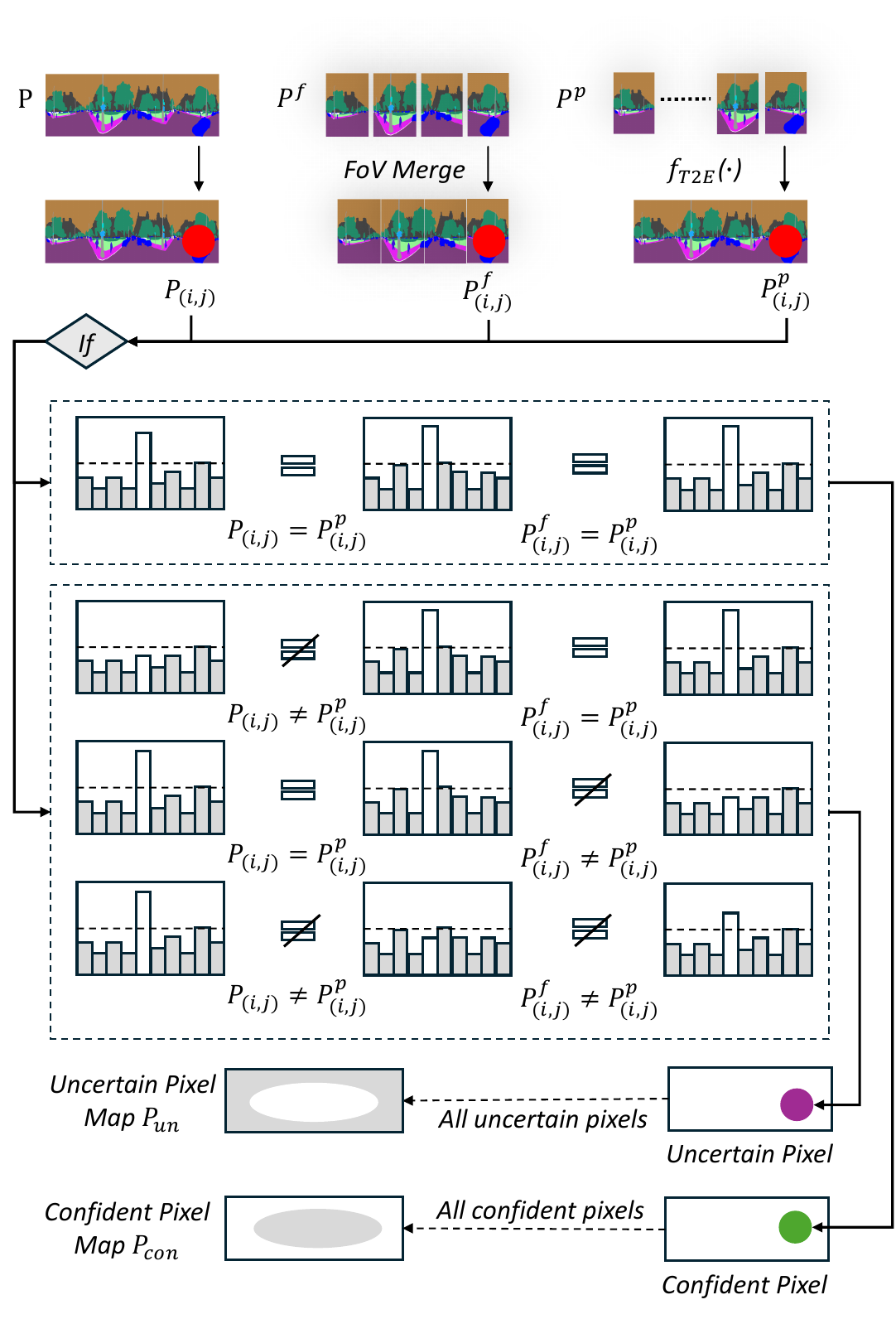}
    \vspace{-20pt}
    \caption{Illustration of the cross-projection pixel-wise prediction assessment.}
    \vspace{-12pt}
    \label{fig:CPPP}
\end{figure}

Compared to prior UDA and SFUDA methods using prototypical adaptation, \eg, CFA~\cite{zheng2023look}, MPA~\cite{zhang2022bending, zhang2022behind} and our previous PPAM~\cite{zheng2024semantics}, our RP$^2$AM possesses two distinct characteristics: 
\textbf{(a)} cross-projection prototypical adaptation is conducted between ERP and the other two projections, \ie, TP and FFP, aiming at alleviating the distortion and semantic mismatch problems; and
\textbf{(b)} intra-projection prototypical adaptation is performed with ERP and FFP, aiming to achieve reliable prototypical adaptation for the ERP and distortion-aware ability for the FFP.
Specifically, we first project the source model predictions $P^p$, $P^f$ into pseudo labels: 
\begin{align}
\label{pseudo}
\setlength{\abovedisplayskip}{3pt}
\setlength{\belowdisplayskip}{3pt}
    &\hat{y}^p_{(h,w,k)} = 1_{k \doteq argmax(P^p_{h,w,:}) }, \qquad \notag \\
    &\hat{y}^f_{(H,W/4,k)} = 1_{k \doteq argmax(P^f_{H,W/4,:}) },
\end{align}
where $k$ denotes the semantic category. Subsequently, we propose the cross-projection pixel-wise prediction assessment to select the confident predicted pixels across three projections.

\subsubsection{Cross-projection Pixel-wise Prediction Assessment}
\label{CPPP}
As shown in Fig.~\ref{fig:CPPP}, we transfer all the prediction maps from different projections into the ERP image (\eg, 400 $\times$ 2048 in DensePASS) by merging the FFP and applying transformation function $f_{T2E}(\cdot)$ as \cite{zheng2023both} to TP images. Then, for each pixel coordinate \((i,j)\), a comparative evaluation of the prediction logits \(P_{(i,j)}\), \(P_{(i,j)}^f\), and \(P_{(i,j)}^p\) is conducted to distinguish confident from uncertain pixel predictions. Specifically, a pixel \((i,j)\) is deemed confident if the condition \(P_{(i,j)} = P_{(i,j)}^f = P_{(i,j)}^p\) holds, signifying concordance in the prediction outputs across the varied projections for $P_{(i,j)}$. This allows for building up a confident pixel map from pixels, such as $P_{(i,j)}$. Conversely, the pixels with different predictions are defined as uncertain pixels, facilitating the generation of an uncertain pixel map. 

\subsubsection{Cross-projection Prototype Extraction}
\label{CPPE}
\begin{figure}[t!]
    \centering
    \includegraphics[width=0.5\textwidth]{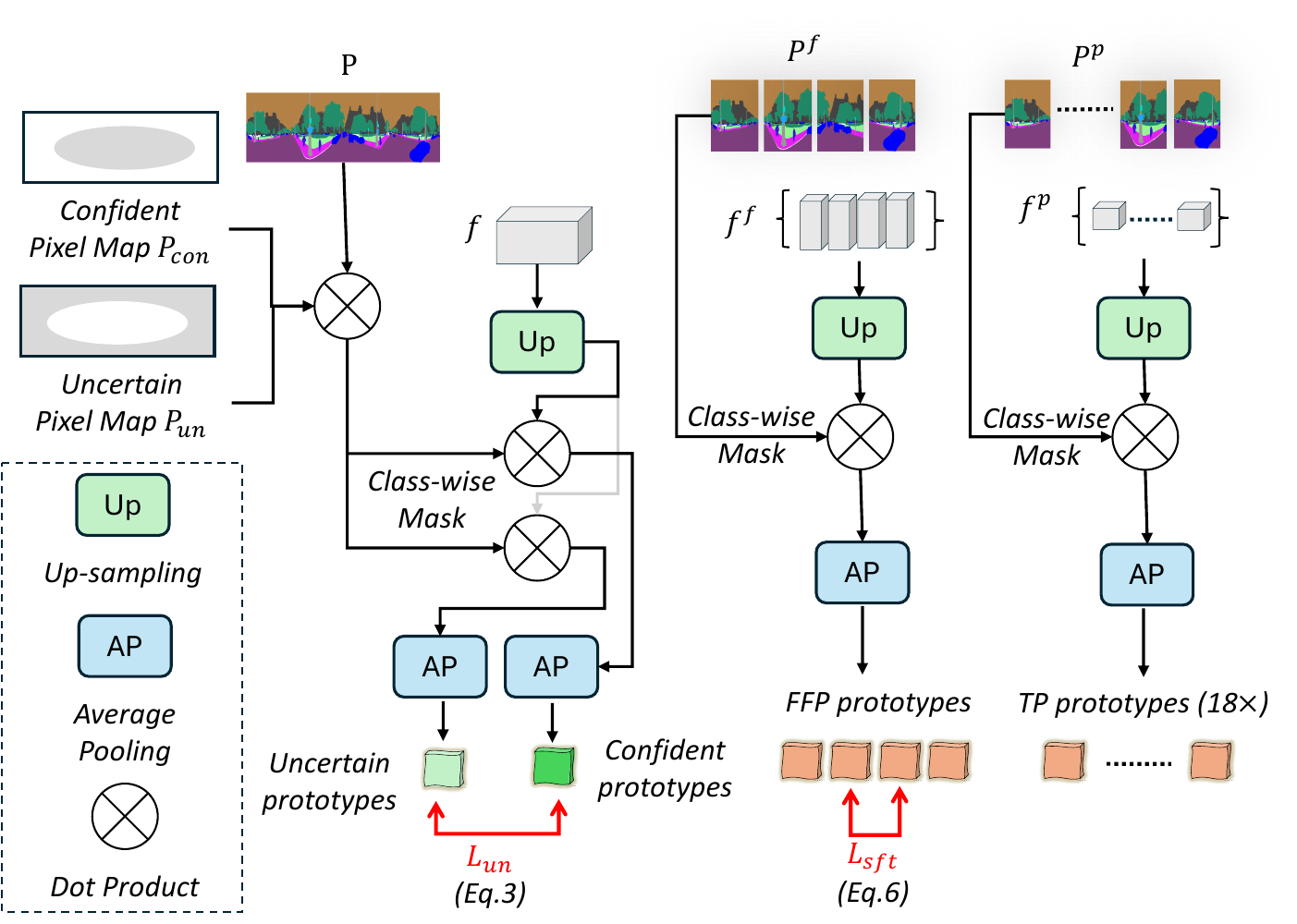}
    \vspace{-12pt}
    \caption{Illustration of the cross-projection prototype extraction. 
    }
    \vspace{-12pt}
    \label{fig:CPE}
\end{figure}

We then utilize the predicted maps as masks to get the confident and uncertain predictions $P_{con}$ and $P_{un}$ which are used to attain the confident and uncertain prototypes for ERP images. $P_{con}$ and $P_{un}$ are first converted to pseudo labels as the same operations in Eq.~\ref{pseudo}. Then we up-sample the ERP features $f$ to fit the spatial size as $P_{con}$ and $P_{un}$. The uncertain and confident prototypes $\tau_{un}$ and $\tau_{con}$ for ERP images are achieved by masked average pooling (MAP) operation, as shown in Fig.~\ref{fig:CPE}. To realize the intro-projection prototypical knowledge adaptation from the reliable pixels to the uncertain pixels, the Mean Squared Error (MSE) loss is imposed between the reliable and uncertain prototypes as follows:
\begin{equation}
    \mathcal{L}_{un} = \sum_{\alpha \neq \beta}^4\{\frac{1}{K}\sum_{k \in K}(\tau_{con} - \tau_{un})^2\}.
\end{equation}

Subsequently, class-specific masked features are derived by amalgamating the up-sampled features with their corresponding pseudo labels, denoted as \(\hat{y}^p_{(h,w,k)}\) for TP and \(\hat{y}^f_{(H,W/4,k)}\) for FFP images. 
It is noteworthy that the prototypes for TP and FFP images, represented as \(\sum_{a=1}^{18} (\tau_p^k)_a\) and \(\sum_{b=1}^4 (\tau_f^k)_b\) respectively, are acquired through the MAP operation, as depicted in Fig.~\ref{fig:CPE}. Within each projection, RP$^2$AM first integrates the prototypes:
\begin{equation}
    \tau_p^k = avg(\sum_{a=1}^{18} (\tau_p^k)_a), \qquad 
    \tau_f^k = avg(\sum_{b=1}^4 (\tau_f^k)_b).
\end{equation}
As shown in Fig.~\ref{fig:overall}, $\tau_p^k$ and $\tau_f^k$ are integrated together as $\tau_{pf}^k$ to retain the less distortion characteristics inherent $\tau_p^k$ and the similar scale semantics of $\tau_f^k$. The $\tau_{pf}^k$ is then used to update the panoramic global prototype $\tau_g^k$, which is iteratively updated with $\tau_{pf}^k$. To obtain more accurate and reliable prototypes, we update $\tau_g^k$ and $\tau_{pf}^k$ as follows:
\begin{equation} 
\label{Eq:iter}
    \tau_g^i = \frac{1}{i}(\tau_{pf}^k)^i + (1 - \frac{1}{i})(\tau_g^k)^{i-1},
\end{equation}
where $(\tau_g^k)^i$ and $(\tau_{pf}^k)^i$ are the prototypes for category $k$ in the $i-$th training epoch, $(\tau_g^k)^{i-1}$ is the panoramic global prototype saved in the last training epoch, $i$ is the current epoch number.
The panoramic global prototype $\tau_g^k$ is then used to give supervision for the confident prototype $\tau_{con}$ obtained from Sec.~\ref{CPPP}, as shown in Eq.~\ref{lossppa}.

\subsubsection{Fine-tuning Source Model for Better Knowledge Extraction}
\label{FSMBKE}
Besides extracting prototypical knowledge from the source model, RP$^2$AM also fine-tunes the source model to improve the effectiveness of knowledge extraction. 
Given that each ERP image is converted to four FFP image patches with the same FoV, the source model consequently extracts four distinct sets of FFP features, denoted as $f_f$. Considering that these features originate from the same ERP image, we synchronize the class-wise prototypes across each set of FFP features, thereby enhancing the source model's overall performance and distortion awareness. Specifically, the prototypes $\sum_{\alpha=1}^4 \tau_{\alpha}$ of the four FFP features are obtained through the same operations with $\tau_g^t$. 

Each FFP image encapsulates a unique, non-overlapping 90$^\circ$ FoV, presenting distinct levels of distortion and overlapping semantic content across the different FFP images. The alignment of prototypes corresponding to each FFP image not only bolsters the source model's capability to recognize and adapt to the distortion but also facilitates the exploration of complementary semantic content inherent within each FFP image patch. 
Thus, the MSE loss is imposed between each two of the prototypes, formulated as follows:
\begin{equation}
    \mathcal{L}_{sft} = \sum_{\alpha \neq \beta}^4\{\frac{1}{K}\sum_{k \in K}((\tau_f^k)_{\alpha}-(\tau_f^k)_{\beta})^2\}.
\end{equation}
Note that $\mathcal{L}_{sft}$ is only used to fine-tune the source model $F_S$.

\subsection{Knowledge Adaptation}
\label{Sec:3.3}
To facilitate the adaptation of knowledge to the target domain, we first enforce loss constraints on both predictions and prototypes. Moreover, we introduce a Cross-Dual Attention Module (\textbf{CDAM}) at the feature level. This module is designed to \textit{enhance the alignment of both spatial and channel characteristics across domains and projections}. The CDAM leverages attention mechanisms~\cite{liu2021source} to adjust and synchronize feature representations, thereby improving the congruence between the source and target domains. This strategic incorporation of spatial and channel attention statistics within CDAM bridges the domain-specific discrepancies, ensuring a more effective and nuanced knowledge transfer.


Specifically, the FFP image patches are stitched to reconstruct an ERP image. It is then passed to the source model $F_S$ to predict a pseudo label, which is employed to supervise the ERP predictions of the target model $F_T$. For simplicity, we use the Cross-Entropy (CE) loss, which is formulated as:
\begin{equation}
    \mathcal{L}_{sup} = CE(P, 1_{k \doteq argmax(\{Rebuild(P^f_{H,W/4,:})\}) }).
\end{equation}
The prototype-level knowledge transfer loss is achieved by MSE loss between the panoramic global prototype $\tau_g^k$ and the confident prototype $\tau_{con}^k$ from ERP images:
\begin{equation}
\label{lossppa}
    \mathcal{L}_{ppa} = \frac{1}{K}\sum_{k \in K}(\tau_g^k-\tau_{con}^k)^2.
\end{equation}
With loss $\mathcal{L}_{ppa}$, the reliable prototypes are pushed together to transfer the knowledge extracted from the source domain to the target domain.
In summary, with the proposed RP$^2$AM, we can effectively address the distortion and semantic mismatch problems at the prediction and prototype levels. We now tackle the style discrepancy problem at the feature level.

\noindent \textbf{Cross-projection Dual Attention Module (CDAM).}
\label{CDAM}
Inspired by the dual attention~\cite{liu2021source}, focusing on spatial and channel characteristics, our CDAM imitates the spatial and channel-wise distributions of cross-projection features to alleviate the style discrepancies in UDA for panoramic semantic segmentation. 
Different from the basic dual attention in \cite{liu2021source} which minimizes the distribution distance of the dual attention maps between the fake source and target data, our CDAM focuses on directly aligning the distribution between FFP and ERP of the same panoramic images rather than introducing additional frameworks or parameters to estimate fake source data as proxy.
As shown in Fig.~\ref{fig:overall}, we reconstruct the FFP features $F^f$ to ensure that the rebuilt feature $F'$ has the same spatial size as $F$. Before the cross dual attention operation, we apply a Batch Normalization Statics (BNS) guided constraint on $F$ and $F'$. Since the BNS of the source model should satisfy the feature distribution of the source data, we align $F$ and $F'$ with BNS to alleviate the domain gaps as follows:
\begin{align}
    \mathcal{L}_{bns} = & {|| \mu(F) - \bar{\mu}||}_2^2 + {|| \sigma^2(F) - \bar{\sigma}^2||}_2^2  \notag \\  
    &+ {|| \mu(F') - \bar{\mu}||}_2^2 + {|| \sigma^2(F') - \bar{\sigma}^2||}_2^2,    
 \end{align}
where $\bar{\mu}$ and $\bar{\sigma}^2$ are the mean and variance parameters of the last BN layer in the source model $S$.

\begin{algorithm}[t!]
\caption{Framework of 360SFUDA++}
\label{alg}
\textbf{Input}: Pre-trained Source Model $F_S$; Unlabeled Panoramic Images (ERP) $x_T$, FFP image patches $x_T^f$, and TP image patches $x_T^t$; \\
\textbf{Output}: Target Model $F_T$; \\
1 Initialize $F_T$ randomly; \\
2 \textbf{for} \textit{each epoch} \textbf{do} \\
3 \hspace{0.5cm} \textbf{for} \textit{each iteration} \textbf{do} \\
4 \hspace{0.5cm} Sample $x_T$ and process $x_T^f$ and $x_T^t$ \\
5 \hspace{0.5cm} Forward Pass of $F_S$ (Eq.1) \\
6 \hspace{0.5cm} Obtain pseudo labels (Eq.2) \\
7 \hspace{0.5cm} Cross-projection prediction assessment (Fig.~\ref{fig:CPPP}) \\
8 \hspace{0.5cm} Prototype extraction (Fig.~\ref{fig:CPE}, Eq.4) \\
9 \hspace{0.5cm} $\mathcal{L}_{un}$ between uncertain and confident prototypes (Eq.3) \\
10\hspace{0.5cm} Update global panoramic prototypes (Eq.5) \\
11\hspace{0.5cm} Prototypical adaptation loss $\mathcal{L}_{ppa}$ (Eq.8) \\
12\hspace{0.5cm} Fine-tuning $F_S$ with $\mathcal{L}_{sft}$ (Eq.6) \\
13\hspace{0.5cm} Align $F$ and $F'$ with $\mathcal{L}_{bns}$ (Eq.9) \\
14\hspace{0.5cm} Feature level knowledge transfer (Eq.12) \\
15\hspace{0.5cm} \textbf{end} \\
16 \textbf{end} \\
17 Final target model $F_T$.
\end{algorithm}

As shown in Fig.~\ref{fig:overall}, after aligned with BNS, the ERP feature $f$ and the rebuilt feature $f'$ are first reshaped to be $f \in \mathbb{R}^{N \times C}$ and $f' \in \mathbb{R}^{N \times C}$, where $N$ is the number of pixels and $C$ is the channel number. Then we calculate the spatial-wise attention maps $M_{sp} \in \mathbb{R}^{N \times C}$ and ${M'}_{sp} \in \mathbb{R}^{N \times C}$ for $f$ and $f'$ by:
\begin{align}
    & \{M_{sp}\}_{ji} = \frac{exp({f'}_{[i:]} \cdot f_{[:j]}^T)}{\sum_i^{N} exp({f'}_{[i:]} \cdot f_{[:j]}^T)}, \qquad \notag \\ 
    & \{M_{sp}'\}_{ji} = \frac{exp(f_{[i:]} \cdot {f'}_{[:j]}^T)}{\sum_i^{N} exp(f_{[i:]} \cdot {f'}_{[:j]}^T)}, 
\end{align}
where $f^T$ is the transpose of $f$ and $\{M\}_{ij}$ measures the impact of the $i$-th position on the $j$-th position. 
Similarly, the channel-wise attention maps $M_{ch} \in \mathbb{R}^{C \times C}$ and ${M'}_{ch} \in \mathbb{R}^{C \times C}$ 
can be obtained through:
\begin{align}
    &\{M_{ch}\}_{ji} = \frac{exp({f'}_{[i:]}^T \cdot f_{[:j]})}{\sum_i^{C} exp({f'}_{[i:]} \cdot f_{[:j]}^T)}, \qquad \notag \\
    &\{M_{ch}'\}_{ji} = \frac{exp(f_{[i:]}^T \cdot {f'}_{[:j]})}{\sum_i^{C} exp(f_{[i:]}^T \cdot {f'}_{[:j]})}.
\end{align}
After obtaining the spatial and channel attention maps, the CDAM loss can be calculated with the Kullback-Liibler divergence (\ie, KL divergence) as follows:
\begin{equation}
    \mathcal{L}_{cda} = KL(M_{sp}, M_{sp}') + KL(M_{ch}, M_{ch}')
\end{equation}

\begin{figure*}[t!]
    \centering\includegraphics[width=\textwidth]{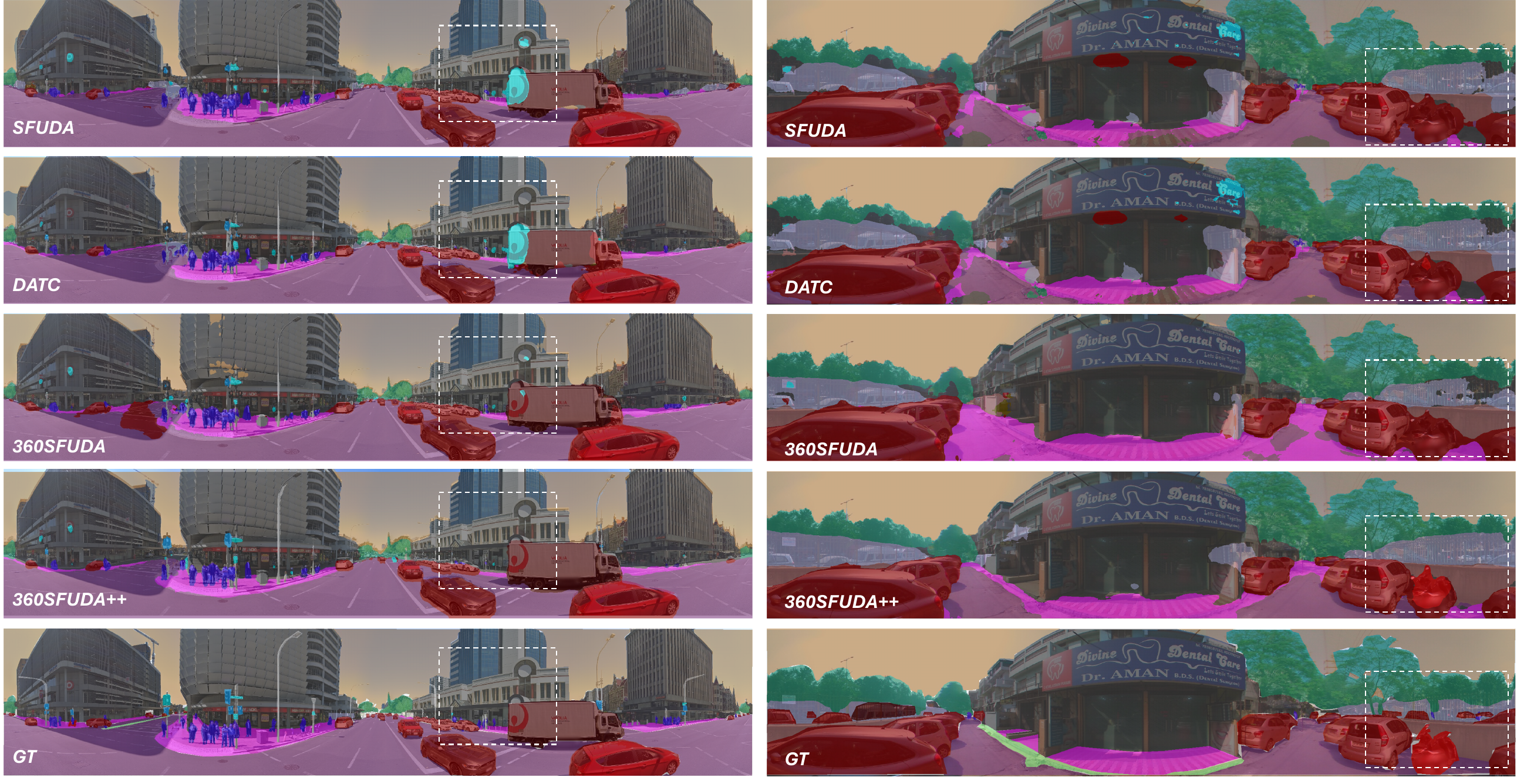}
    \vspace{-16pt}
    \caption{Visualization results on C-to-D scenario. (a) source, (b) SFDA~\cite{liu2021source}, (c) DATC~\cite{yang2022source}, (d) 360SFUDA, (e) 360SFUDA++ (f) Ground Truth (GT).}
    \label{fig:C2D}
    \vspace{-6pt}
\end{figure*}

\begin{table*}[t!]
\centering
\setlength{\tabcolsep}{1pt}
\caption{Experimental results of 19 selected categories in panoramic semantic segmentation on C-to-D. SF: Source-free UDA. The \textbf{bold} and \underline{underline} denote the best and the second-best performance in source-free UDA methods, respectively.}
\resizebox{0.99\textwidth}{!}{
\begin{tabular}{l|c|c|cccccccccccccccccccc}
\toprule 
Method & SF & mIoU & Road & S.W. & Build. & Wall & Fence & Pole & Light & Sign & Vegt. & Terr. & Sky & Pers. & Rider & Car & Truck & Bus & Train & Motor & Bike  \\ \midrule 
ECANet~\cite{yang2021capturing} (O.s.) & \XSolidBrush & 43.02 & 81.60 & 19.46 & 81.00 & 32.02 & 39.47 & 25.54 & 3.85 & 17.38 & 79.01 & 39.75 & 94.60 & 46.39 & 12.98 & 81.96 & 49.25 & 28.29 & 0.00 & 55.36 & 29.47  \\ 
P2PDA~\cite{p2pda} & \XSolidBrush & 41.99 & 70.21 & 30.24 & 78.44 & 26.72 & 28.44 & 14.02 & 11.67 & 5.79 & 68.54 & 38.20 & 85.97 & 28.14 & 0.00 & 70.36 & 60.49 & 38.90 & 77.80 & 39.85 & 24.02 \\ 
SIM~\cite{wang2020differential} (S.t.) & \XSolidBrush & 44.58 & 68.16 & 32.59 & 80.58 & 25.68 & 31.38 & 23.60 & 19.39 & 14.09 & 72.65 & 26.41 & 87.88 & 41.74 & 16.09 & 73.56 & 47.08 & 42.81 & 56.35 & 47.72 & 39.30 \\ 
PCS~\cite{PCS} & \XSolidBrush & 53.83 & 78.10 & 46.24 & 86.24 & 30.33 & 45.78 & 34.04 & 22.74 & 13.00 & 79.98 & 33.07 & 93.44 & 47.69 & 22.53 & 79.20 & 61.59 & 67.09 & 83.26 & 58.68 & 39.80 \\ 
DAFormer~\cite{hoyer2022daformer} & \XSolidBrush & 54.67 & 73.75 & 27.34 & 86.35 & 35.88 & 45.56 & 36.28 & 25.53 & 10.65 & 79.87 & 41.64 & 94.74 & 49.69 & 25.15 & 77.70 & 63.06 & 65.61 & 86.68 & 65.12 & 48.13 \\ 
Trans4PASS-T ~\cite{zhang2022bending} & \XSolidBrush & 53.18 & 78.13 & 41.19    & 85.93    & 29.88 & 37.02  & 32.54 & 21.59         & 18.94        & 78.67      &45.20   &93.88 & 48.54  & 16.91 & 79.58 & 65.33 & 55.76 & 84.63 & 59.05      & 37.61   \\ 
DPPASS-T~\cite{zheng2023both} & \XSolidBrush &55.30 &78.74 &46.29 &87.47 &48.62 &40.47 &35.38 &24.97 &17.39 &79.23 &40.85 &93.49 &52.09 &29.40 &79.19 &58.73 &47.24 &86.48 &66.60 &38.11 \\ 
DATR-S~\cite{zheng2023look} & \XSolidBrush &56.81 &80.63 &51.77 &87.80 &44.94 &43.73 &37.23 &25.66 &21.00& 78.61 &26.68 &93.77 &54.62 &29.50 &80.03 &67.35 &63.75 &87.67 &67.57 &37.10  \\ 
\midrule
Source &  \Checkmark & 38.65 & 65.26 & 29.40 & 77.04 & 15.14 & 28.72 & 14.15 & 9.36 & 10.55 & 69.09 & 21.10 & 82.91 & 40.98 & 10.42 & 68.56 & 32.90 & 44.94 & 50.98 & 37.74 & 25.19 \\
SFDA~\cite{liu2021source} & \Checkmark & 42.70 & 68.75 & 31.59 & 80.99 & 19.61 & 29.60 & 18.67 & 7.7 & 14.08 & 73.74 & 24.91 & 88.38 & 41.66 & 8.46 & 69.97 & 47.48 & 33.25 & 72.02 & 47.62 & 32.77 \\
DATC~\cite{yang2022source} & \Checkmark & 43.06 & 70.21 & 35.87 & 80.60 & 21.42 & 28.14 & 19.10 & 5.79 & 15.10 & 72.76 & \underline{27.42} & 88.14 & 41.65 & 10.29 & 72.32 & 47.80 & 21.97 & 80.91 & 46.65 & 32.01 \\
SFUDA~\cite{tian2024self} & \Checkmark & 43.40 & 68.64 & 44.55 & 79.30 & 32.44 & 30.48 & 21.70 & 13.35 & 16.84 & 75.26 & 24.74 & 86.70 & 38.80 & 8.97 & 75.39 & 40.90 & 4.99 & \textbf{85.24} & 48.28 & 27.94 \\
360SFUDA~\cite{zheng2024semantics} w/ b1 & \Checkmark & 48.78 & \underline{72.90} & \textbf{48.10} & 82.77 & 22.19 & \underline{39.69} & 26.66 & 17.90 & 14.35 & 74.98 & 25.95 & 88.95 & 45.36 & 15.83 & 75.70 & 49.16 & 55.68 & 82.07 & 54.82 & 33.76  \\ 
360SFUDA~\cite{zheng2024semantics} w/ b2 & \Checkmark & 50.12 & 72.29 & 43.04 & \underline{84.48} & \underline{29.72} & 37.68 & 22.83 & 9.52 & 14.45 & \underline{75.26} & \textbf{34.53} & 91.12 & 49.92 & \underline{27.22} & 76.22 & 47.81 & \textbf{64.13} & 79.47 & 56.83 & \textbf{35.76} \\
\rowcolor{gray!10} 360SFUDA++ w/ b1 & \Checkmark & \underline{50.19} & 68.59 & 46.70 & 84.06 & 29.08 & 34.74 & \textbf{31.28} & \underline{20.31} & \underline{16.84} & 75.04 & 23.07 & \underline{92.20} & \underline{50.03} & 18.32 & \underline{78.75} & \textbf{56.53} & 49.90 & \underline{83.63} & \underline{59.00} & \underline{35.46} \\ 
\rowcolor{gray!20} 360SFUDA++ w/ b2 & \Checkmark & \textbf{52.99} & \textbf{74.00} & \underline{48.03} & \textbf{85.86} & \textbf{34.41} & \textbf{41.28} & \underline{31.25} & \textbf{22.59} & \textbf{17.41} & \textbf{76.74} & 26.24 & \textbf{92.65} & \textbf{54.60} & \textbf{30.76} & \textbf{79.41} & \underline{55.60} & \underline{59.84} & 81.25 & \textbf{59.89} & 35.02 \\
\bottomrule
\end{tabular}}
\vspace{-5pt}
\label{Tab:City2PASS}
\end{table*}

\subsection{Optimization}
The overall optimization procedure is shown in Algorithm.~\ref{alg}. The training objective for learning the target model containing three losses is defined as:
\begin{equation}
    \mathcal{L} = \mathcal{L}_{un} + \mathcal{L}_{ppa} + \gamma \cdot \mathcal{L}_{cda} + \mathcal{L}_{bns} + \mathcal{L}_{sup}
\end{equation}
where $\mathcal{L}_{ppa}$ is the MSE loss from RP$^2$AM, $\mathcal{L}_{cda}$ refers to the KL loss from CDAM, $\mathcal{L}_{sup}$ denotes the CE loss for the prediction pseudo label supervision loss, $\mathcal{L}_{bns}$ refers to the BNS guided feature loss, and $\lambda$ and $\gamma$ are the trade-off weights of the proposed loss terms. 

\begin{figure*}[t!]
    \centering\includegraphics[width=\textwidth]{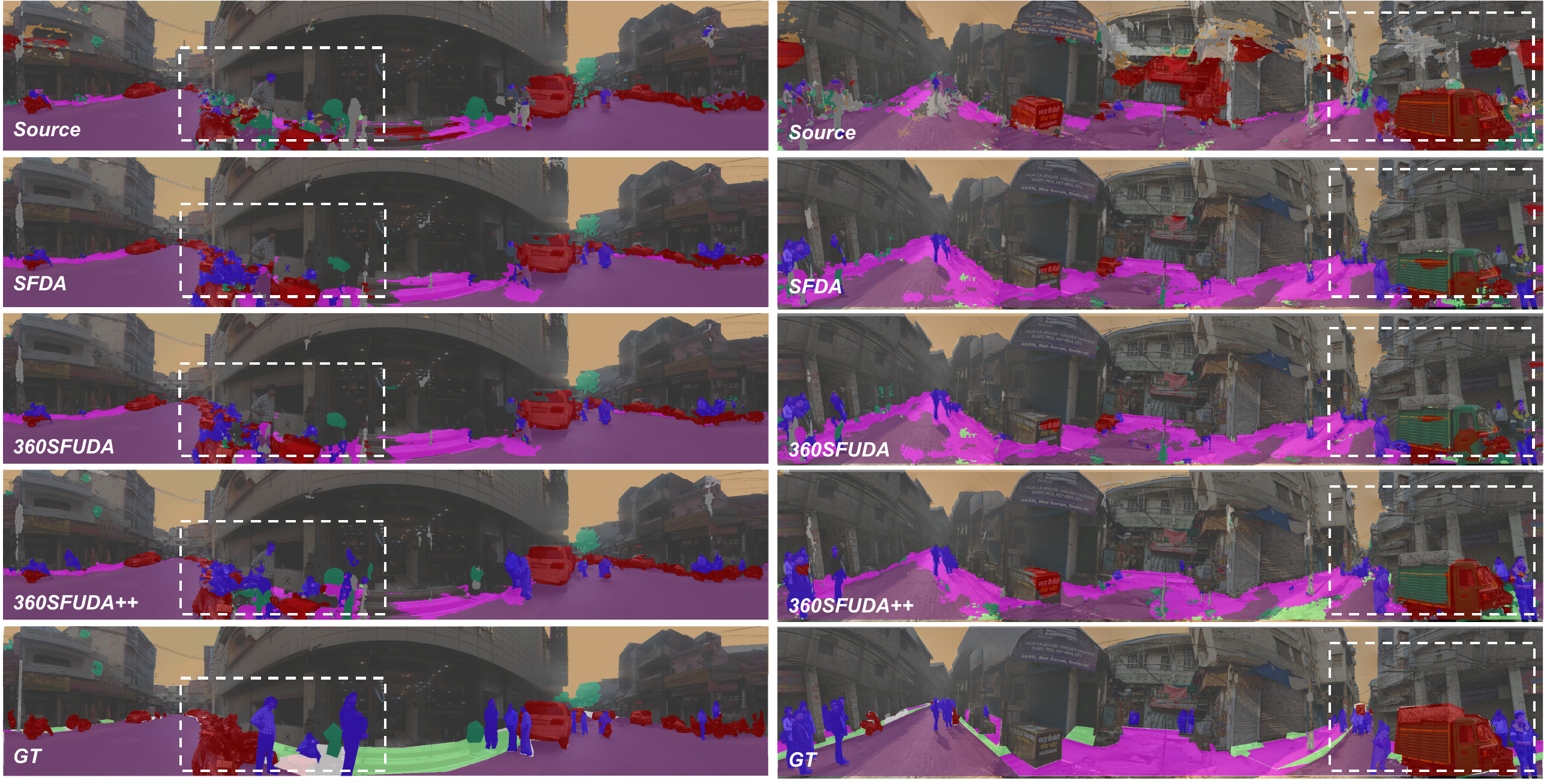}
    \vspace{-16pt}
    \caption{Visualization on Stanford2D3D dataset. (a) RGB panoramic images; (b) 360SFUDA~\cite{zheng2024semantics}; (c) 360SFUDA++; and (d) Ground Truth (GT).}
    \label{fig:dp13_vis}
\end{figure*}
\begin{table*}[h!]
\setlength{\tabcolsep}{2.8pt}
\caption{Experimental results on the S-to-D scenario, the overlapped 13 classes of two datasets are used to test the UDA performance. The \textbf{bold} and \underline{underline} denote the best and the second-best performance in source-free UDA methods, respectively.}
\resizebox{0.99\textwidth}{!}{
\begin{tabular}{l|c|c|ccccccccccccc|c}
\toprule
Method & SF & mIoU & Road & S.W. & Build. & Wall & Fence & Pole & Tr.L. & Tr.S. & Veget. & Terr. & Sky & Pers. & Car & $\Delta$ \\ \midrule
PVT~\cite{wang2021pyramid} SSL & \XSolidBrush & 38.74 & 55.39 & 36.87 & 80.84 & 19.72 & 15.18 & 8.04 & 5.39 & 2.17 & 72.91 & 32.01 & 90.81 & 26.76 & 57.40 &-\\
PVT~\cite{wang2021pyramid} w/ MPA & \XSolidBrush & 40.90 & 70.78 & 42.47 & 82.13 & 22.79 & 10.74 & 13.54 & 1.27 & 0.30 & 71.15 & 33.03 & 89.69 & 29.07 & 64.73 &-\\
Trans4PASS~\cite{zhang2022behind} w/ SSL & \XSolidBrush & 43.17 & 73.72 & 43.31 & 79.88 & 19.29 & 16.07 & 20.02 & 8.83 & 1.72 & 67.84 & 31.06 & 86.05 & 44.77 & 68.58 &-\\
Trans4PASS~\cite{zhang2022behind} w/ MPA & \XSolidBrush & 45.29 & 67.28 & 43.48 & 83.18 & 22.02 & 21.98 & 22.72 & 7.86 & 1.52 & 73.12 & 40.65 & 91.36 & 42.69 & 70.87 &-\\ 
DATR-M~\cite{zheng2023look} w/ CFA& \XSolidBrush & 51.04 & 77.62 & 49.03 & 84.58 & 28.15 & 39.70 & 30.34 & 23.83 & 15.95 & 78.23 & 24.74 & 93.73 & 44.14 & 74.45 & - \\
\midrule
Source w/ seg-b1 & \Checkmark & 35.81 & 63.36 & 24.09 & 80.13 & \textbf{15.68} & 13.39 & 16.26 & 7.42 & 0.09 & 62.45 & 20.20 & 86.05 & 23.02 & 53.37 &- \\
SFDA w/ seg-b1~\cite{liu2021source} & \Checkmark & 38.21 & 68.78 & 30.71 & 80.37 & 5.26 & 18.95 & 20.90 & 5.25 & 2.36 & 70.19 & 23.30 & 90.20 & 22.55 & 57.90 &+2.40 \\
ProDA w/ seg-b1~\cite{zhang2021prototypical} & \Checkmark & 37.37 & 68.93 & 30.88 & 80.07 & 4.17 & 18.60 & 19.72 & 1.77 & 1.56 & 70.05 & 22.73 & 90.60 & 19.71 & 57.04 & +2.73 \\
GTA w/ seg-b1~\cite{kundu2021generalize} & \Checkmark & 36.00& 64.61 & 20.04 & 79.04 & 8.06 & 15.36 & 19.86 & 6.02 & 2.13 & 65.77 & 17.75 & 84.56 & 26.71 & 58.13 & +0.19 \\
HCL w/ seg-b1~\cite{huang2021model} & \Checkmark & 38.38 & 68.82 & 30.41 & 80.37 & 5.88 & 20.18 & 20.10 & 4.23 & 2.11 & 70.50 & 24.74 & 89.89 & 22.65 & 59.04 & +2.57 \\
DATC w/ seg-b1~\cite{yang2022source} & \Checkmark & 38.54 & 69.48 & 26.96 & 80.68 & 11.64 & 15.24 & 20.10 & 9.33 & 0.55 & 66.11 & 24.31 & 85.16 & 30.90 & 60.58 &+2.73\\ 
Simt w/ seg-b1~\cite{guo2022simt} & \Checkmark & 37.94 & 68.47 & 29.51 & 79.62 & 6.78 & 19.20 & 19.48 & 2.31 & 1.33 & 68.85 & 26.55 & 89.30 & 22.35 & 59.49 & +2.13 \\ 
SFUDA~\cite{tian2024self} & \Checkmark & 41.68 & 72.05 & 37.27 & 81.88 & 14.11 & 24.96 & 23.73 & 0.00 & 0.39 & 70.55 & 33.19 & 90.51 & 29.99 & 63.17 & +5.87 \\
360SFUDA~\cite{zheng2024semantics} w/ b1 & \Checkmark & 41.78 & \underline{70.17} & 33.24 & 81.66 & 13.06 & 23.40 & 23.37 & 7.63 & \underline{3.59} & 71.04 & 25.46 & 89.33 & \underline{36.60} & 64.60 & +5.97\\ 
360SFUDA~\cite{zheng2024semantics} w/ b2 & \Checkmark & 42.18 & 69.99 & 32.28 & 81.34 & 10.62 & 24.35 & 24.29 & \textbf{9.19} & \textbf{3.63} & 71.28 & \underline{30.04} & 88.75 & \textbf{37.49} & \underline{65.05} & +6.37\\ 
\rowcolor{gray!10} 360SFUDA++ w/ b1 & \Checkmark & 42.53 & 69.65 & \underline{33.49} & \underline{81.90} & 11.35 & \underline{24.80} & 24.90 & 8.92 & 2.14 & \underline{71.79} & 26.58 & \textbf{90.65} & 35.49 & 64.32 & \underline{+6.72} \\ 
\rowcolor{gray!20} 360SFUDA++ w/ b2 & \Checkmark & \textbf{43.43} & \textbf{71.06} & \textbf{34.95} & \textbf{82.61} & \underline{14.23} & \textbf{27.82} & \textbf{25.05} & \underline{9.18} & 2.82 & \textbf{72.81} & \textbf{32.34} & \underline{90.39} & 36.20 & \textbf{65.13} & \textbf{+7.62} \\ 
\bottomrule
\end{tabular}}
\vspace{-12pt}
\label{Tab:syn2dense}
\end{table*}

\begin{figure*}[t!]
    \centering\includegraphics[width=\textwidth]{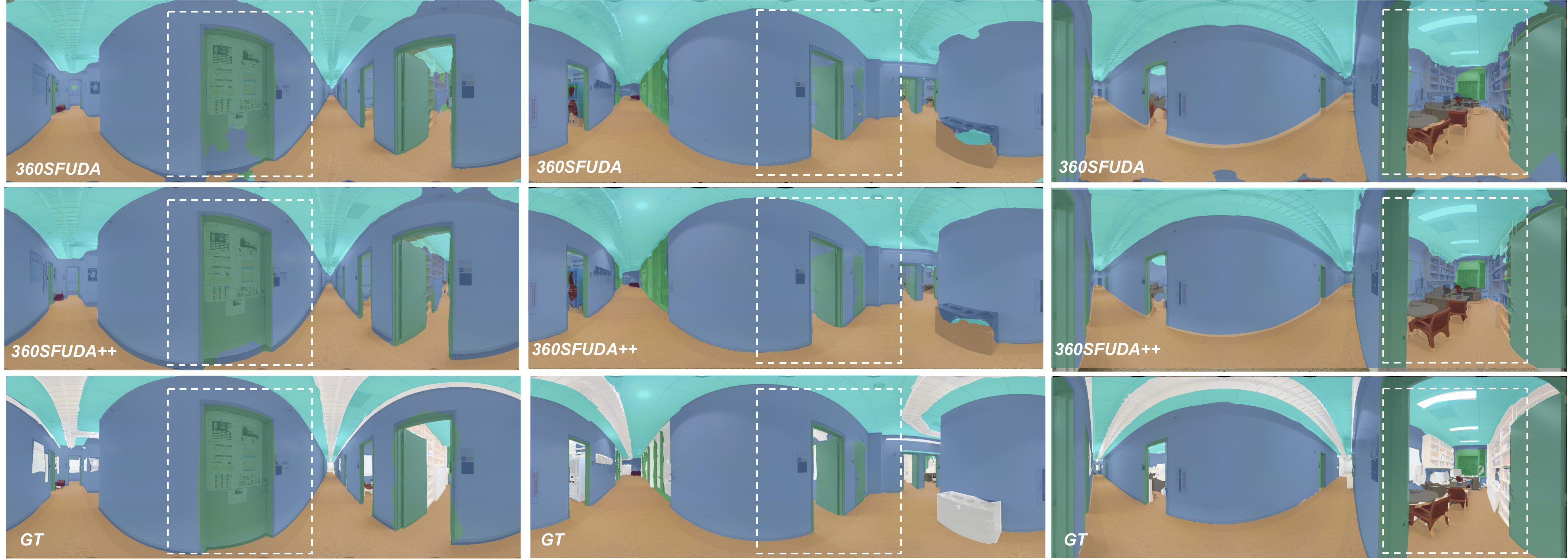}
    \vspace{-20pt}
    \caption{Visualization on Stanford2D3D dataset. (a) RGB panoramic images; (b) 360SFUDA~\cite{zheng2024semantics}; (c) 360SFUDA++; and (d) Ground Truth (GT).}
    \vspace{-8pt}
    \label{fig:stanford_vis}
\end{figure*}

\begin{table*}[t!]
\centering
\label{tab:stanford}
\caption{Experimental results on indoor Stanford2D3D~\cite{armeni2017joint}. The \textbf{bold} denotes the best performance among UDA and SFUDA methods.}
\setlength{\tabcolsep}{7pt}
\resizebox{0.99\linewidth}{!}{
\begin{tabular}{l|c|c|cccccccc|c}
\toprule
Method & SF & mIoU & Ceiling & Chair & Door & Floor & Sofa & Table & Wall & Window & $\Delta$ \\ \midrule
PVT-S~\cite{zhang2022bending} & \XSolidBrush & 57.71 & 85.69 & 51.71 & 18.54 & 90.78 & 34.76 & 65.34 & 74.87 & 39.98 & - \\
PVT-S w/ MPA~\cite{zhang2022bending} & \XSolidBrush & 57.95 & 85.85 & 51.76 & 18.39 & 90.78 & 35.93 & 65.43 & 75.00 & 40.43 & - \\
Trans4PASS w/ MPA~\cite{zhang2022bending} & \XSolidBrush & 64.52 & 85.08 & 58.72 & 34.97 & 91.12 & 46.25 & 71.72 & 77.58 & 50.75 & - \\
Trans4PASS+~\cite{zhang2022behind} & \XSolidBrush & 63.73 & 90.63 & 62.30 & 24.79 & 92.62 & 35.73 & 73.16 & 78.74 & 51.78 & - \\
Trans4PASS+ w/ MPA~\cite{zhang2022behind} & \XSolidBrush & 67.16 & 90.04 & 64.04 & 42.89 & 91.74 & 38.34 & 71.45 & 81.24 & 57.54 & -
\\ \midrule
SFDA~\cite{liu2021source}  & \Checkmark & 54.76 & 79.44 & 33.20 & 52.09 & 67.36 & 22.54 & 53.64 & 69.38 & 60.46 & -\\
360SFUDA~\cite{zheng2024semantics} w/ b1 & \Checkmark & 57.63 & 73.81 & 29.98 & \underline{63.65} & 73.49 & 31.76 & 49.25 & 72.89 & 66.22 & +2.87\\ 
360SFUDA~\cite{zheng2024semantics} w/ b2 & \Checkmark & 65.75 & 82.88 & 38.00 & \textbf{65.81} & 86.71 & 36.32 & \underline{66.10} & \textbf{80.29} & \underline{69.88} & +10.99\\ 
\rowcolor{gray!10} \textbf{360SFUDA++} w/ b1 & \Checkmark & \underline{66.54} & \textbf{86.28} & \textbf{58.25} & 36.04 & \textbf{87.99} & \underline{48.74} & 65.44 & 77.82 & \textbf{71.78} & \underline{+11.78} \\ 
\rowcolor{gray!20} \textbf{360SFUDA++} w/ b2 & \Checkmark & \textbf{68.84} & \underline{85.50} &\underline{57.59} & 53.15 & \underline{87.40} & \textbf{53.63} & \textbf{66.49} & \underline{80.23} & 66.75 & \textbf{+14.08} \\
\bottomrule
\end{tabular}}
\vspace{-5pt}
\end{table*}

\section{Experiments and Analysis}
\label{Experiments and Analysis}
We empirically validate our method by comparing it with the existing SFUDA, UDA, and panoramic segmentation methods on three widely used benchmarks.

\subsection{Datasets and Implementation Details.}

\noindent (1) Cityscapes-to-DensePASS \textbf{(C-to-D):} Cityscapes~\cite{Cityscapes} constitutes a comprehensive real-world dataset meticulously collected for autonomous driving applications, encompassing street scenes sourced from 50 distinct cities. It offers precise pixel-wise annotations across 19 semantic categories. The dataset's official split includes 2975 images for training and 500 images for validation. In this study, we leverage the official training set (2975 images) as the source domain data for acquiring the pre-trained source model. DensePASS~\cite{densepass}, on the other hand, is a panoramic dataset tailored to capture diverse street scenes worldwide, comprising 2500 panoramas. Within this collection, a subset of 100 panoramas has been meticulously annotated with a high degree of precision, targeting categories that are crucial for navigation. This subset is designated as the test set for evaluation purposes. In the context of C-to-D scenario, we employ the DensePASS training set as the unlabeled target panoramic data.

\noindent (2) SynPASS-to-DensePASS \textbf{(S-to-D):} SynPASS~\cite{zhang2022behind} emerges as a synthetic dataset composed of 9080 synthetic panoramic images meticulously annotated across 22 categories. Its official sets for training, validation, and testing encompass 5700, 1690, and 1690 images, respectively. For training and testing purposes, we concentrate on the subset of 13 categories that SynPASS and DensePASS datasets have in common for both training and evaluation purposes. This setup allows us to explore the nuances of transferring knowledge from synthetic (SynPASS) to real-world data (DensePASS), treating SynPASS as the synthetic source dataset and DensePASS as the real-world target dataset.

\noindent (3) Stanford2D3D-pinhole-to-panoramic \textbf{(SPin-to-SPan):} The Stanford2D3D Pinhole (SPin) dataset comprises 70496 pinhole images annotated across 13 semantic categories. Conversely, the Stanford2D3D Panoramic (SPan) dataset consists of 1413 indoor panoramic images. These images are preciesely annotated with identical 13 semantic categories as observed in the pinhole dataset~\cite{armeni2017joint}. In this paper, our focus narrows to the 8 semantic categories that are common across both subsets.

\noindent \textbf{Implementation Details.} We train our frameworks with 4 NVIDIA GPUs with an initial learning rate of 6e$^{-5}$, which is scheduled by the poly strategy with power 0.9 over 200 epochs. The optimizer is AdamW with epsilon 1e$^{-8}$, weight decay 1e$^{-4}$, and batch size is 4 on each GPU. The adaptation frameworks are trained within 10K iterations. With respect to input resolutions, training images from indoor pinhole cameras and panoramic views are processed at resolutions of 1080 $\times$ 1080 and 1024 $\times$ 512, respectively. In the SynPASS-to-DensePASS (S-to-D) scenario, synthetic panoramic images maintain a resolution of 1024 $\times$ 512. For outdoor scenarios, pinhole and synthetic panoramic images are adjusted to 1024 $\times$ 512, whereas real-world panoramic images are presented at 2048 $\times$ 400. The resolutions for validation image sets in indoor and outdoor environments are set to 2048 $\times$ 1024 and 2048 $\times$ 400, respectively, to ensure uniformity in the evaluation framework.

\begin{figure*}[t!]
    \centering\includegraphics[width=0.99\textwidth]{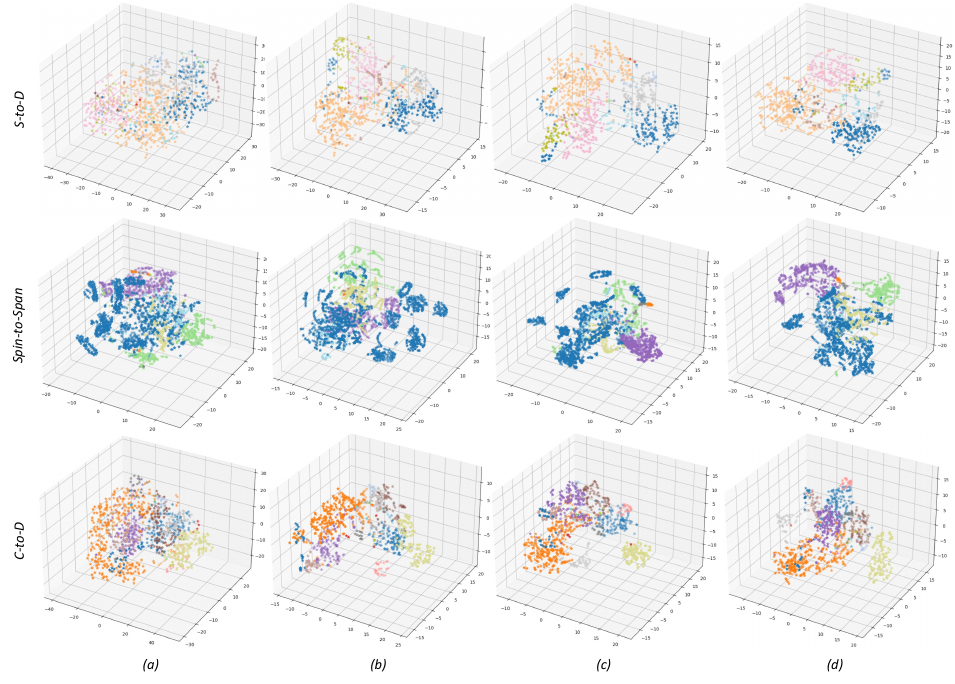}
    \vspace{-12pt}
    \caption{TSNE Visualization on C-to-D and Spin-to-Span scenarios. (a) source, (b) SFDA~\cite{liu2021source}, (c) 360SFUDA~\cite{zheng2024semantics}, and (d) 360SFUDA++.}
    \vspace{-12pt}
    \label{fig:tsne}
\end{figure*}

\subsection{Experimental Results.}
\label{Experimental Results}
Initially, we undertake a comprehensive evaluation of the proposed 360SFUDA++ framework within the C-to-D scenario. Outcomes, including both qualitative and quantitative analyses, are presented in Fig.~\ref{fig:C2D} and Tab.~\ref{Tab:City2PASS}, demonstrating the framework's adaptation capabilities across urban panoramas. 360SFUDA++ demonstrates superior performance relative to the previous version 360SFUDA and other source-free UDA methods~\cite{liu2021source, yang2022source}. Remarkably, it also achieves panoramic semantic segmentation performance that are comparable to those of UDA methods that leverage source domain imagery during the adaptation process, such as Trans4PASS~\cite{zhang2022behind} (52.99\% vs. 53.18\%). Our 360SFUDA++ model yields substantial improvements, with mIoU increases of +1.41\% and +2.87\% using SegFormer-B1 and -B2 backbones, surpassing our previous version, 360SFUDA, and significantly outperforming SFUDA methods SFDA~\cite{liu2021source} and DATC~\cite{yang2022source} by +7.49\% and +7.13\% in mIoU, respectively. The qualitative results are shown in the Fig.~\ref{fig:C2D}, vividly illustrating our 360SFUDA++'s superior capability in handling complex segmentation tasks.
We also provide the TSNE visualization to further show the superiority of our 360SFUDA++ in Fig.~\ref{fig:tsne}, apparently, our 360SFUDA++ gains a significant improvement in distinguishing the pixels in panoramic images in both prediction and the high-level representation spaces.

Subsequently, we conduct an evaluation of our 360SFUDA++ framework within the S-to-D scenario. The outcomes of this assessment are detailed in  Tab.~\ref{Tab:syn2dense} and Fig.~\ref{fig:dp13_vis}. 
Apparently, our 360SFUDA++ significantly outperforms the source-free UDA methods~\cite{liu2021source,yang2022source} by a large margin and 360SFUDA++ improves the previous version~\cite{zheng2024semantics} by +0.75 and +1.25 mIoU with SegFormer-B1 and -B2 backbones, respectively. 
The results affirm that our 360SFUDA++ framework, augmented by RP$^2$AM and CDAM, is particularly adept at facilitating robust and efficacious knowledge transfer across domains for panoramic semantic segmentation. Moreover, the performance of 360SFUDA++ in key categories pertinent to driving—such as ‘person' (54.60\% mIoU), ‘rider' (30.76\% mIoU), and ‘car' (79.41\% mIoU)—exemplifies its superior capability in semantic segmentation within critical traffic-related contexts. Additionally, the t-SNE visualization presented in Figure~\ref{fig:tsne} further elucidates the effectiveness of 360SFUDA++ in cultivating a distinctive and informative representation space, underscoring its overall proficiency in domain adaptation for panoramic imaging tasks.
 
Lastly, we evaluate our 360SFUDA++ on the Spin-to-Span scenario can compare it with our previous 360SFUDA~\cite{zheng2024semantics} and SFDA~\cite{liu2021source} methods, as well as the UDA methods, such as MPA~\cite{zhang2022behind}. As illustrated in Tab.~\ref{tab:stanford}, 360SFUDA++ significantly outperforms the source-free method SFDA~\cite{liu2021source} by +11.78\% mIoU. Notably, 360SFUDA++ even outperforms the UDA methods using source data in the adaptation (68.84\% vs. 67.16\% mIoU). The visualization results in Fig.~\ref{fig:stanford_vis} also demonstrate the superiority of our 360SFUDA++. Additionally, t-SNE visualizations highlight 360SFUDA++'s enhanced capability for distinguishing features within high-level representation spaces, reaffirming its effectiveness in domain adaptation for panoramic semantic segmentation. \textit{\textbf{More quantitative comparison and visualization results refer to the supplementary material.}}

\begin{table}[]
\setlength{\tabcolsep}{2pt}
\caption{Ablation study of different loss Combinations.}
\resizebox{0.48\textwidth}{!}{
\begin{tabular}{cccccccccc}
\toprule
\multicolumn{6}{c}{Loss Combinations} & \multicolumn{2}{c}{C-to-D} & \multicolumn{2}{c}{S-to-D} \\ \midrule
$\mathcal{L}_{sup}$ & $\mathcal{L}_{un}$ & $\mathcal{L}_{ppa}$ & $\mathcal{L}_{sft}$ & $\mathcal{L}_{cda}$ & $\mathcal{L}_{bns}$ & mIoU & $\Delta$ & mIoU & $\Delta$ \\ \midrule
\Checkmark &  &  &  &  &  & 38.65 & - & 35.81 & - \\ 
\Checkmark & \Checkmark &  &  &  &  & 43.05 & +4.40 & 36.44 & +0.63 \\ 
\Checkmark & \Checkmark & \Checkmark &  &  &  & 51.05 & +12.40 & 40.96 & +5.15 \\ 
\Checkmark & \Checkmark & \Checkmark & \Checkmark &  &  & 51.16 & +12.51 & 41.80 & +5.99 \\
\Checkmark &  &  &  & \Checkmark &  & 44.24 & +5.59 & 38.38 & +2.57 \\ 
\Checkmark &  &  &  & \Checkmark & \Checkmark & 44.79 & +6.14 & 38.52 & +2.71 \\ 
\Checkmark & \Checkmark & \Checkmark & \Checkmark & \Checkmark & \Checkmark & 52.99 & +14.34 & 42.53 & +6.72 \\ \bottomrule
\end{tabular}
\vspace{-20pt}
\label{LossCombin}
}
\end{table}

\section{Ablation Study}
\label{Ablation Study}
\subsection{Different Loss Function Combinations.}
Tab.~\ref{LossCombin} evaluates the influences of all the proposed loss functions in 360SFUDA++, including $\mathcal{L}_{sup}$, $\mathcal{L}_{un}$, $\mathcal{L}_{ppa}$, $\mathcal{L}_{sft}$, $\mathcal{L}_{cda}$, and $\mathcal{L}_{bns}$. Building upon the foundational loss functions introduced in~\cite{zheng2024semantics}, 360SFUDA++ innovatively integrates $\mathcal{L}{un}$ to facilitate local prototypical knowledge transfer from reliable to uncertain pixels within target panoramic imagery. Moreover, it refines $\mathcal{L}{ppa}$ through the application of MSE between the global panoramic prototype and the confident prototype extracted from ERP images. As delineated in Tab.~\ref{LossCombin}, all the proposed loss functions have a positive impact on improving the overall panoramic semantic segmentation performance. 
Noteworthy is the RP$^2$AM, leveraging $\mathcal{L}{un}$ and $\mathcal{L}{ppa}$, which secures substantial improvements of +12.40\% and +5.15\% in mIoU, underscoring the module's efficacy in optimizing knowledge extraction and adaptation processes. Furthermore, the CDAM exerts a favorable influence on segmentation outcomes, evidencing increments of +5.59\% and +2.57\% mIoU across C-to-D and S-to-D scenarios, respectively.
 These results affirm CDAM's success in replicating spatial and channel-level feature distributions between ERP and FFP images, effectively mitigating stylistic disparities across domains.

\subsection{Ablation Study of RP$^2$AM}
\label{Sec:ab}
\begin{figure}[t!]
    \centering
    \includegraphics[width=0.5\textwidth]{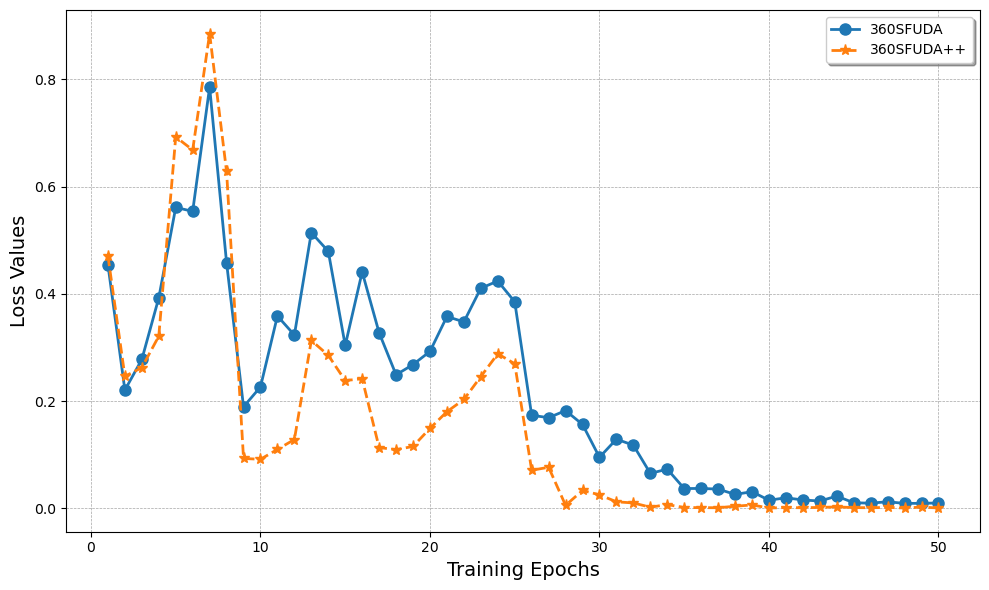}
    \vspace{-16pt}
    \caption{Loss curves of the prototypical adaptation methods on C-to-D scenario, \ie, PPAM and RP$^2$AM, respectively.}
    \label{fig:loss_compare}
    \vspace{-8pt}
\end{figure}

\noindent \textbf{PPAM vs. RP$^2$AM.}
In order to ascertain the efficacy of the proposed RP$^2$AM in comparison to PPAM delineated in our previous work ~\cite{zheng2024semantics}, we conduct ablation experiments and provide the visualization of loss curves of PPAM and RP$^2$AM on the C-to-D scenario. The results are shown in Fig.~\ref{fig:loss_compare}. The graphical representation clearly illustrates that, although RP$^2$AM's loss trajectory experiences more pronounced fluctuations during the initial stages of training, it eventually achieves a more favorable convergence compared to PPAM throughout the training duration. This observation underscores RP$^2$AM's superior capacity for stable and effective learning, marking a significant advancement over PPAM.

\begin{figure*}[t!]
    \centering
    \includegraphics[width=\linewidth]{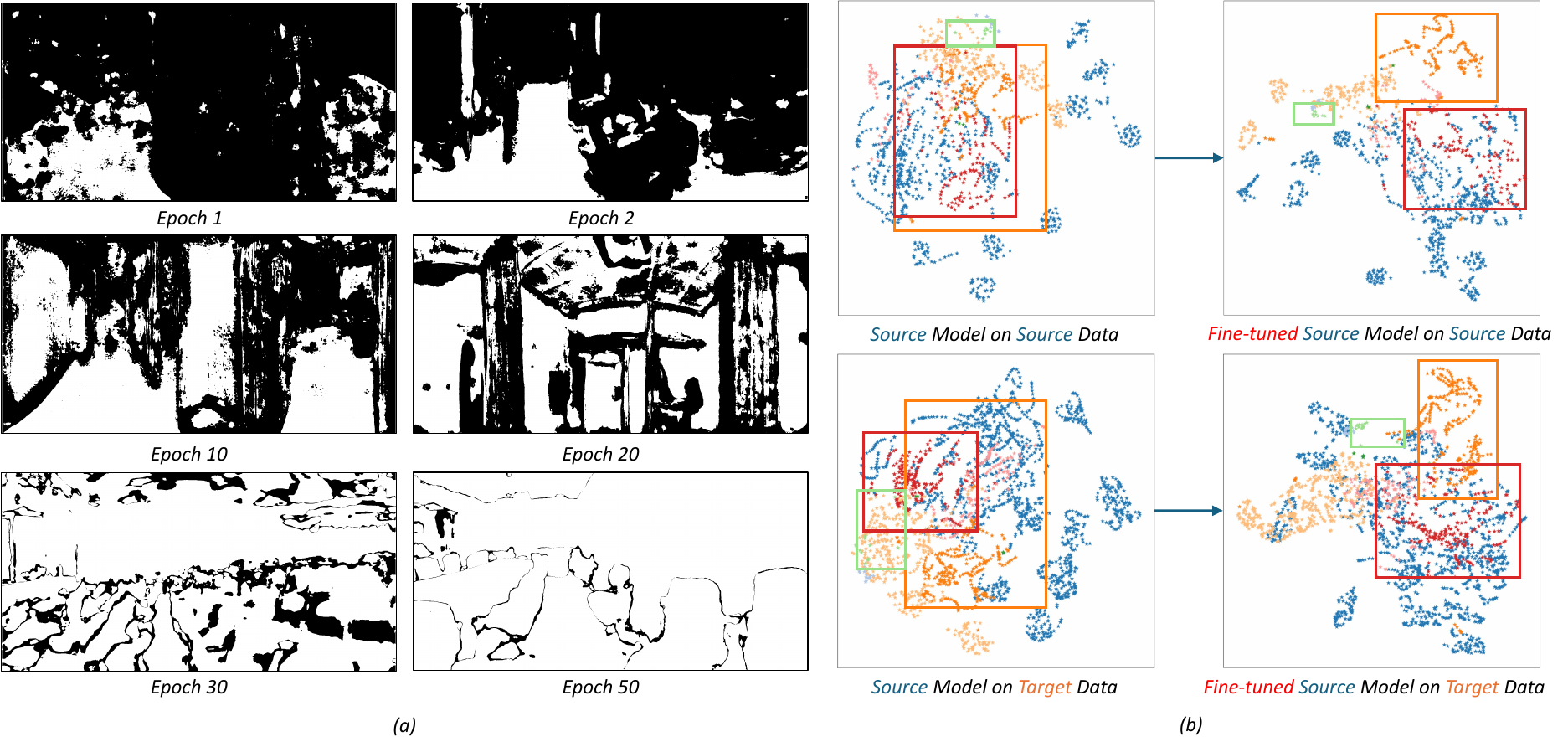}
    \vspace{-24pt}
    \caption{(a) Confidence maps across the training procedure, white pixels represent the confident predictions while black pixels show the uncertain predictions. (b) TSNE visualization of fine-tuning source model with $\mathcal{L}_{sft}$. 
    }
    \vspace{-12pt}
    \label{fig:ablation_fig}
\end{figure*}

\noindent \textbf{Effectiveness of the Confidence Maps}
To substantiate the efficacy of the cross-projection pixel-wise prediction assessment employed within the RP$^2$AM, we present visualizations of confidence maps throughout the training process, as illustrated in Fig.~\ref{fig:ablation_fig} (a). These visualizations reveal a progressive increase in the prevalence of white pixels within the confidence maps, symbolizing a growth in the number of pixels deemed confident. Concurrently, quantitative analyses, as detailed in Tab.~\ref{relia_ab}, corroborate that the implementation of reliable prototypical adaptation in 360SFUDA++ leads to considerable performance enhancements over the framework's preceding iteration detailed in~\cite{zheng2024semantics}. This convergence of qualitative and quantitative evidence firmly establishes the beneficial impact of employing confidence maps for improved pixel-wise prediction accuracy in panoramic semantic segmentation tasks.

\begin{table}[]
\centering
\setlength{\tabcolsep}{8pt}
\caption{Ablation of RP$^2$AM vs. PPAM on C-to-D scenario.}
\resizebox{\linewidth}{!}{
\begin{tabular}{ccc}
\toprule
Combinations & PPAM ($\mathcal{L}_{ppa}$ in ~\cite{zheng2024semantics})  & RP$^2$AM ($\mathcal{L}_{un}$+$\mathcal{L}_{ppa}$) \\ \midrule
mIoU & 45.42 & \textbf{51.05}    \\ \bottomrule
\end{tabular}}
\label{relia_ab}
\vspace{-12pt}
\end{table}

\subsection{Ablation Study of CDAM}
\noindent\textbf{Dual Attention vs. Cross Dual Attention.}
The dual attention (DA) approach proposed in SFDA~\cite{liu2021source} aligns the spatial and channel characteristics of features between the fake source and target data. In contrast, our cross dual attention (CDA) approach aligns the distribution between different projections of the same spherical data, specifically ERP and FFP, resulting in more robust and stable knowledge transfer. Moreover, in our CDA, we obtain spatial and channel characteristics across features from different projections, whereas DA operates within features. We compare DA and our cross-projection DA on the C-to-D scenario, our method achieves 44.24\% mIoU, while DA only reaches 41.53\% mIoU. This empirical evidence suggests that our cross-projection DA offers a more adept solution for addressing the challenges inherent in SFUDA for panoramic semantic segmentation tasks.

\noindent \textbf{FoV Selection of FFP.}
The integration of distortion-aware mechanisms and innovative projection techniques has been extensively investigated within the scope of panoramic semantic segmentation in existing literature, such as~\cite{zhang2022bending, zhang2022behind} and Zheng et al.~\cite{zheng2023both}. Nonetheless, as delineated in Sec.~\ref{Sec:intro}, the complex semantic content and object relationships inherent in 360-degree imagery precipitate a pronounced semantic incongruity across different domains. Intuitively, we introduce the Fixed FoV Projection (FFP) strategy to address the aforementioned problem. The quantitative results in Tab.~\ref{FFPFoV} show that the FoV of 90$^\circ$ achieves the best panoramic segmentation performance of 44.28\% mIoU. This finding underscores the effectiveness of the FFP strategy in mitigating domain-specific semantic discrepancies in panoramic domain.


\begin{table}[t!]
\setlength{\tabcolsep}{6pt}
\caption{Ablation study of the FoV of our proposed FFP.}
\resizebox{0.49\textwidth}{!}{
\begin{tabular}{cccccccc}
\toprule
FoV & w/o & $60^\circ$ & $72^\circ$ & $90^\circ$ & $120^\circ$ & $180^\circ$ & $360^\circ$ \\ \midrule
mIoU & 38.65 & 44.03 & 44.16 & \textbf{44.28} & 44.02 & 41.65 & 40.31 \\ 
$\Delta$ & - & +5.38 & +5.51 & \textbf{+5.63} & +5.37 & +3.00 & +1.66 \\
\bottomrule
\end{tabular}}
\label{FFPFoV}
\vspace{-12pt}
\end{table}

\begin{table}[t!]
\centering
\caption{Ablation study of the trade-off parameters $\gamma$ for $\mathcal{L}_{cda}$.}
\resizebox{0.49\textwidth}{!}{
\setlength{\tabcolsep}{8pt}
\begin{tabular}{ccccccc}
\toprule
$\gamma$ & 0 & 0.01 & 0.02 & 0.05 & 0.1 & 0.2 \\ \midrule
        mIoU & 38.65 & 42.05 & 43.24 & 43.28 & \textbf{44.24} & 43.07 \\ 
        $\Delta$ & - & +3.40 & +4.59 & +4.63 & \textbf{+5.59} & +4.42 \\
\bottomrule \\
\end{tabular}}
\label{Tab:gammaab}
\vspace{-12pt}
\end{table}



\noindent \textbf{Ablation of $\gamma$.}
We study the sensitivity of hyper-parameter $\gamma$, which are the trade-off weights for the KL loss in CDAM. The quantitative results are provided in Tab.~\ref{Tab:gammaab}.

\section{Discussion}
\noindent \textbf{Fine-tuning the Source Model.}
In the context of domain adaptation, the process of transferring knowledge from a pre-trained source model to a target domain presents notable challenges. The intrinsic discrepancies between the source pinhole domain and the target panoramic image domain necessitate a tailored approach for model adaptation. To this end, we advocate for the refinement of the pre-trained source model utilizing a specialized fine-tuning loss, denoted as $\mathcal{L}_{sft}$. This is predicated on the observation that a model pre-trained in the source domain may not inherently exhibit optimal performance within the target domain.
 
The results presented in Tab.~\ref{LossCombin} substantiates the efficacy of the proposed fine-tuning loss $\mathcal{L}_{sft}$ in enhancing model performance. Further insight is afforded by the TSNE visualizations depicted in Fig.~\ref{fig:ablation_fig} (b), which contrast the performance of the source model before and after fine-tuning on both source and target domain data. These visualizations compellingly illustrate that fine-tuning with unlabeled target data not only elevates the model's effectiveness within the target domain but also augments its capacity for pixel-wise feature differentiation in the source domain. This nuanced approach underscores the importance of fine-tuning pre-trained models with domain-specific adaptations to bridge the gap between disparate domains, thereby improving model generalizability and performance across different modalities.

\noindent \textbf{Difference between Pinhole-to-Panoramic and Pinhole-to-Pinhole UDA.}
In the context of SFUDA, the primary challenge arises from the domain shift attributable to style discrepancies, which stem from differing data distribution between the source and target domains. However, when the objective is to adapt models from the conventional pinhole camera domain to the domain of panoramic images, one encounters additional complexities. As evidenced by the results in Tba.~\ref{Tab:City2PASS}, endeavors have been made for adaptation within the pinhole camera domain~\cite{liu2021source, yang2022source} tend to offer marginal improvements in performance when bechmarked against a standard baseline. This phenomenon is largely due to the more intricate nature of domain shift that occur in the transition from pinhole to panoramic adaptation scenarios, a topic elaborated upon in Sec.~\ref{Sec:intro}. By integrating our proposed modules, \ie, RP$^2$AM and CDAM, the performance has been largely improved by +6.08\% and +5.72\% performance gain over SFDA~\cite{liu2021source} and DATC~\cite{yang2022source}, respectively.


\section{Conclusion and Future Work}
In this paper, we explored the challenging source-free unsupervised domain adaptation from pinhole to panoramic image domain. To this end, we proposed a novel framework for panoramic semantic segmentation, namely 360SFUDA++, to address the domain shifts between pinhole and panoramic images, including semantic mismatch, inevitable distortion, and inherent style discrepancies. We subtly utilized the multi-projection versatility of 360$^\circ$ data for efficient and reliable domain knowledge transfer. The RP$^2$AM is introduced to select the confident extracted knowledge and integrate panoramic prototypes for reliable knowledge adaptation. Meanwhile, we proposed CDAM to better align the spatial and channel characteristics across projections at the feature level between domains. Extensive experiments on the synthetic and real-world benchmarks, including indoor and outdoor scenarios, show that our 360SFUDA++ outperforms prior approaches and is on par with the UDA methods using the source data.

\noindent\textbf{Future Work:}
In future research, we intend to employ Large Language Models (LLMs) and Multi-modal Large Language Models (MLLMs) to address the domain discrepancies, specifically targeting semantic mismatches between pinhole and panoramic images. This approach aims to enhance the interpretative alignment and integration across the different projection of spherical data. Also, the Segment Anything Model (SAM) can be applied to unleash its powerful zero-shot instance segmentation ability.

{
    \small
    \bibliographystyle{IEEEtran}
    \bibliography{main}

\begin{thebibliography}{10}
\providecommand{\url}[1]{#1}
\csname url@samestyle\endcsname
\providecommand{\newblock}{\relax}
\providecommand{\bibinfo}[2]{#2}
\providecommand{\BIBentrySTDinterwordspacing}{\spaceskip=0pt\relax}
\providecommand{\BIBentryALTinterwordstretchfactor}{4}
\providecommand{\BIBentryALTinterwordspacing}{\spaceskip=\fontdimen2\font plus
\BIBentryALTinterwordstretchfactor\fontdimen3\font minus
  \fontdimen4\font\relax}
\providecommand{\BIBforeignlanguage}[2]{{%
\expandafter\ifx\csname l@#1\endcsname\relax
\typeout{** WARNING: IEEEtran.bst: No hyphenation pattern has been}%
\typeout{** loaded for the language `#1'. Using the pattern for}%
\typeout{** the default language instead.}%
\else
\language=\csname l@#1\endcsname
\fi
#2}}
\providecommand{\BIBdecl}{\relax}
\BIBdecl

\bibitem{360survey}
H.~Ai, Z.~Cao, J.~Zhu, H.~Bai, Y.~Chen, and L.~Wang, ``Deep learning for
  omnidirectional vision: A survey and new perspectives,'' \emph{arXiv preprint
  arXiv:2205.10468}, 2022.

\bibitem{yang2019pass}
K.~Yang, X.~Hu, L.~M. Bergasa, E.~Romera, and K.~Wang, ``Pass: Panoramic
  annular semantic segmentation,'' \emph{IEEE Transactions on Intelligent
  Transportation Systems}, vol.~21, no.~10, pp. 4171--4185, 2019.

\bibitem{yang2020omnisupervised}
K.~Yang, X.~Hu, Y.~Fang, K.~Wang, and R.~Stiefelhagen, ``Omnisupervised
  omnidirectional semantic segmentation,'' \emph{IEEE Transactions on
  Intelligent Transportation Systems}, 2020.

\bibitem{zhang2022bending}
J.~Zhang, K.~Yang, C.~Ma, S.~Rei{\ss}, K.~Peng, and R.~Stiefelhagen, ``Bending
  reality: Distortion-aware transformers for adapting to panoramic semantic
  segmentation,'' in \emph{Proceedings of the IEEE/CVF Conference on Computer
  Vision and Pattern Recognition}, 2022, pp. 16\,917--16\,927.

\bibitem{p2pda}
J.~Zhang, C.~Ma, K.~Yang, A.~Roitberg, K.~Peng, and R.~Stiefelhagen, ``Transfer
  beyond the field of view: Dense panoramic semantic segmentation via
  unsupervised domain adaptation,'' \emph{IEEE Transactions on Intelligent
  Transportation Systems}, 2021.

\bibitem{PCS}
X.~Yue, Z.~Zheng, S.~Zhang, Y.~Gao, T.~Darrell, K.~Keutzer, and A.~S.
  Vincentelli, ``Prototypical cross-domain self-supervised learning for
  few-shot unsupervised domain adaptation,'' in \emph{Proceedings of the
  IEEE/CVF Conference on Computer Vision and Pattern Recognition}, 2021, pp.
  13\,834--13\,844.

\bibitem{zheng2024semantics}
X.~Zheng, P.~Zhou, A.~Vasilakos, and L.~Wang, ``Semantics, distortion, and
  style matter: Towards source-free uda for panoramic segmentation,'' 2024.

\bibitem{zheng2023both}
X.~Zheng, J.~Zhu, Y.~Liu, Z.~Cao, C.~Fu, and L.~Wang, ``Both style and
  distortion matter: Dual-path unsupervised domain adaptation for panoramic
  semantic segmentation,'' \emph{arXiv preprint arXiv:2303.14360}, 2023.

\bibitem{zhang2022behind}
J.~Zhang, K.~Yang, H.~Shi, S.~Rei{\ss}, K.~Peng, C.~Ma, H.~Fu, K.~Wang, and
  R.~Stiefelhagen, ``Behind every domain there is a shift: Adapting
  distortion-aware vision transformers for panoramic semantic segmentation,''
  \emph{arXiv preprint arXiv:2207.11860}, 2022.

\bibitem{kirillov2023segment}
A.~Kirillov, E.~Mintun, N.~Ravi, H.~Mao, C.~Rolland, L.~Gustafson, T.~Xiao,
  S.~Whitehead, A.~C. Berg, W.-Y. Lo \emph{et~al.}, ``Segment anything,''
  \emph{arXiv preprint arXiv:2304.02643}, 2023.

\bibitem{liu2021source}
Y.~Liu, W.~Zhang, and J.~Wang, ``Source-free domain adaptation for semantic
  segmentation,'' in \emph{Proceedings of the IEEE/CVF Conference on Computer
  Vision and Pattern Recognition}, 2021, pp. 1215--1224.

\bibitem{ye2021source}
M.~Ye, J.~Zhang, J.~Ouyang, and D.~Yuan, ``Source data-free unsupervised domain
  adaptation for semantic segmentation,'' in \emph{Proceedings of the 29th ACM
  International Conference on Multimedia}, 2021, pp. 2233--2242.

\bibitem{yang2022source}
C.-Y. Yang, Y.-J. Kuo, and C.-T. Hsu, ``Source free domain adaptation for
  semantic segmentation via distribution transfer and adaptive class-balanced
  self-training,'' in \emph{2022 IEEE International Conference on Multimedia
  and Expo (ICME)}.\hskip 1em plus 0.5em minus 0.4em\relax IEEE, 2022, pp.
  1--6.

\bibitem{zhang2021prototypical}
P.~Zhang, B.~Zhang, T.~Zhang, D.~Chen, Y.~Wang, and F.~Wen, ``Prototypical
  pseudo label denoising and target structure learning for domain adaptive
  semantic segmentation,'' in \emph{Proceedings of the IEEE/CVF conference on
  computer vision and pattern recognition}, 2021, pp. 12\,414--12\,424.

\bibitem{kundu2021generalize}
J.~N. Kundu, A.~Kulkarni, A.~Singh, V.~Jampani, and R.~V. Babu, ``Generalize
  then adapt: Source-free domain adaptive semantic segmentation,'' in
  \emph{Proceedings of the IEEE/CVF International Conference on Computer
  Vision}, 2021, pp. 7046--7056.

\bibitem{huang2021model}
J.~Huang, D.~Guan, A.~Xiao, and S.~Lu, ``Model adaptation: Historical
  contrastive learning for unsupervised domain adaptation without source
  data,'' \emph{Advances in Neural Information Processing Systems}, vol.~34,
  pp. 3635--3649, 2021.

\bibitem{guo2022simt}
X.~Guo, J.~Liu, T.~Liu, and Y.~Yuan, ``Simt: Handling open-set noise for domain
  adaptive semantic segmentation,'' in \emph{Proceedings of the IEEE/CVF
  Conference on Computer Vision and Pattern Recognition}, 2022, pp. 7032--7041.

\bibitem{zheng2023look}
X.~Zheng, T.~Pan, Y.~Luo, and L.~Wang, ``Look at the neighbor: Distortion-aware
  unsupervised domain adaptation for panoramic semantic segmentation,'' in
  \emph{Proceedings of the IEEE/CVF International Conference on Computer
  Vision}, 2023, pp. 18\,687--18\,698.

\bibitem{zhang2024goodsam}
W.~Zhang, Y.~Liu, X.~Zheng, and L.~Wang, ``Goodsam: Bridging domain and
  capacity gaps via segment anything model for distortion-aware panoramic
  semantic segmentation,'' \emph{arXiv preprint arXiv:2403.16370}, 2024.

\bibitem{zhang2019category}
Q.~Zhang, J.~Zhang, W.~Liu, and D.~Tao, ``Category anchor-guided unsupervised
  domain adaptation for semantic segmentation,'' \emph{Advances in neural
  information processing systems}, vol.~32, 2019.

\bibitem{chen2019domain}
M.~Chen, H.~Xue, and D.~Cai, ``Domain adaptation for semantic segmentation with
  maximum squares loss,'' in \emph{Proceedings of the IEEE/CVF International
  Conference on Computer Vision}, 2019, pp. 2090--2099.

\bibitem{hoyer2022daformer}
L.~Hoyer, D.~Dai, and L.~Van~Gool, ``Daformer: Improving network architectures
  and training strategies for domain-adaptive semantic segmentation,'' in
  \emph{Proceedings of the IEEE/CVF Conference on Computer Vision and Pattern
  Recognition}, 2022, pp. 9924--9935.

\bibitem{vu2019dada}
T.-H. Vu, H.~Jain, M.~Bucher, M.~Cord, and P.~P{\'e}rez, ``Dada: Depth-aware
  domain adaptation in semantic segmentation,'' in \emph{Proceedings of the
  IEEE/CVF International Conference on Computer Vision}, 2019, pp. 7364--7373.

\bibitem{zou2018unsupervised}
Y.~Zou, Z.~Yu, B.~Kumar, and J.~Wang, ``Unsupervised domain adaptation for
  semantic segmentation via class-balanced self-training,'' in
  \emph{Proceedings of the European conference on computer vision (ECCV)},
  2018, pp. 289--305.

\bibitem{stan2021unsupervised}
S.~Stan and M.~Rostami, ``Unsupervised model adaptation for continual semantic
  segmentation,'' in \emph{Proceedings of the AAAI conference on artificial
  intelligence}, vol.~35, no.~3, 2021, pp. 2593--2601.

\bibitem{fleuret2021uncertainty}
F.~Fleuret \emph{et~al.}, ``Uncertainty reduction for model adaptation in
  semantic segmentation,'' in \emph{Proceedings of the IEEE/CVF Conference on
  Computer Vision and Pattern Recognition}, 2021, pp. 9613--9623.

\bibitem{shen2021unsupervised}
W.~Shen, Q.~Wang, H.~Jiang, S.~Li, and J.~Yin, ``Unsupervised domain adaptation
  for semantic segmentation via self-supervision,'' in \emph{2021 IEEE
  International Geoscience and Remote Sensing Symposium IGARSS}.\hskip 1em plus
  0.5em minus 0.4em\relax IEEE, 2021, pp. 2747--2750.

\bibitem{vu2019advent}
T.-H. Vu, H.~Jain, M.~Bucher, M.~Cord, and P.~P{\'e}rez, ``Advent: Adversarial
  entropy minimization for domain adaptation in semantic segmentation,'' in
  \emph{Proceedings of the IEEE/CVF Conference on Computer Vision and Pattern
  Recognition}, 2019, pp. 2517--2526.

\bibitem{pan2020unsupervised}
F.~Pan, I.~Shin, F.~Rameau, S.~Lee, and I.~S. Kweon, ``Unsupervised
  intra-domain adaptation for semantic segmentation through self-supervision,''
  in \emph{Proceedings of the IEEE/CVF Conference on Computer Vision and
  Pattern Recognition}, 2020, pp. 3764--3773.

\bibitem{araslanov2021self}
N.~Araslanov and S.~Roth, ``Self-supervised augmentation consistency for
  adapting semantic segmentation,'' in \emph{Proceedings of the IEEE/CVF
  Conference on Computer Vision and Pattern Recognition}, 2021, pp.
  15\,384--15\,394.

\bibitem{zheng2022uncertainty}
X.~Zheng, C.~Fu, H.~Xie, J.~Chen, X.~Wang, and C.-W. Sham, ``Uncertainty-aware
  deep co-training for semi-supervised medical image segmentation,''
  \emph{Computers in Biology and Medicine}, vol. 149, p. 106051, 2022.

\bibitem{chen2022uncertainty}
J.~Chen, C.~Fu, H.~Xie, X.~Zheng, R.~Geng, and C.-W. Sham, ``Uncertainty
  teacher with dense focal loss for semi-supervised medical image
  segmentation,'' \emph{Computers in Biology and Medicine}, vol. 149, p.
  106034, 2022.

\bibitem{zhu2023good}
J.~Zhu, Y.~Luo, X.~Zheng, H.~Wang, and L.~Wang, ``A good student is cooperative
  and reliable: Cnn-transformer collaborative learning for semantic
  segmentation,'' in \emph{Proceedings of the IEEE/CVF International Conference
  on Computer Vision}, 2023, pp. 11\,720--11\,730.

\bibitem{zheng2022transformer}
X.~Zheng, Y.~Luo, H.~Wang, C.~Fu, and L.~Wang, ``Transformer-cnn cohort:
  Semi-supervised semantic segmentation by the best of both students,''
  \emph{arXiv preprint arXiv:2209.02178}, 2022.

\bibitem{chen4617170frozen}
J.~Chen, D.~Deguchi, C.~Zhang, X.~Zheng, and H.~Murase, ``Frozen is better than
  learning: A new design of prototype-based classifier for semantic
  segmentation,'' \emph{Available at SSRN 4617170}.

\bibitem{zheng2023distilling}
X.~Zheng, Y.~Luo, P.~Zhou, and L.~Wang, ``Distilling efficient vision
  transformers from cnns for semantic segmentation,'' \emph{arXiv preprint
  arXiv:2310.07265}, 2023.

\bibitem{chen2023clip}
J.~Chen, D.~Deguchi, C.~Zhang, X.~Zheng, and H.~Murase, ``Clip is also a good
  teacher: A new learning framework for inductive zero-shot semantic
  segmentation,'' \emph{arXiv preprint arXiv:2310.02296}, 2023.

\bibitem{xie2023adversarial}
H.~Xie, C.~Fu, X.~Zheng, Y.~Zheng, C.-W. Sham, and X.~Wang, ``Adversarial
  co-training for semantic segmentation over medical images,'' \emph{Computers
  in biology and medicine}, vol. 157, p. 106736, 2023.

\bibitem{zheng2024eventdance}
X.~Zheng and L.~Wang, ``Eventdance: Unsupervised source-free cross-modal
  adaptation for event-based object recognition,'' \emph{arXiv preprint
  arXiv:2403.14082}, 2024.

\bibitem{yeh2021sofa}
H.-W. Yeh, B.~Yang, P.~C. Yuen, and T.~Harada, ``Sofa: Source-data-free feature
  alignment for unsupervised domain adaptation,'' in \emph{Proceedings of the
  IEEE/CVF Winter Conference on Applications of Computer Vision}, 2021, pp.
  474--483.

\bibitem{zhao2022source}
Y.~Zhao, Z.~Zhong, Z.~Luo, G.~H. Lee, and N.~Sebe, ``Source-free open compound
  domain adaptation in semantic segmentation,'' \emph{IEEE Transactions on
  Circuits and Systems for Video Technology}, vol.~32, no.~10, pp. 7019--7032,
  2022.

\bibitem{bateson2022source}
M.~Bateson, H.~Kervadec, J.~Dolz, H.~Lombaert, and I.~B. Ayed, ``Source-free
  domain adaptation for image segmentation,'' \emph{Medical Image Analysis},
  vol.~82, p. 102617, 2022.

\bibitem{tian2024self}
Y.~Tian, J.~Li, H.~Fu, L.~Zhu, L.~Yu, and L.~Wan, ``Self-mining the confident
  prototypes for source-free unsupervised domain adaptation in image
  segmentation,'' \emph{IEEE Transactions on Multimedia}, 2024.

\bibitem{sun2024you}
Z.~Sun, L.~Lin, and Y.~Yu, ``You only label once: A self-adaptive
  clustering-based method for source-free active domain adaptation,'' \emph{IET
  Image Processing}, 2024.

\bibitem{Hoffman2018CyCADACA}
J.~Hoffman, E.~Tzeng, T.~Park, J.-Y. Zhu, P.~Isola, K.~Saenko, A.~A. Efros, and
  T.~Darrell, ``Cycada: Cycle-consistent adversarial domain adaptation,'' in
  \emph{ICML}, 2018.

\bibitem{Choi2019SelfEnsemblingWG}
J.~Choi, T.~Kim, and C.~Kim, ``Self-ensembling with gan-based data augmentation
  for domain adaptation in semantic segmentation,'' \emph{2019 IEEE/CVF
  International Conference on Computer Vision (ICCV)}, pp. 6829--6839, 2019.

\bibitem{Sankaranarayanan2018LearningFS}
S.~Sankaranarayanan, Y.~Balaji, A.~Jain, S.-N. Lim, and R.~Chellappa,
  ``Learning from synthetic data: Addressing domain shift for semantic
  segmentation,'' \emph{2018 IEEE/CVF Conference on Computer Vision and Pattern
  Recognition}, pp. 3752--3761, 2018.

\bibitem{Tsai2018LearningTA}
Y.-H. Tsai, W.-C. Hung, S.~Schulter, K.~Sohn, M.-H. Yang, and M.~Chandraker,
  ``Learning to adapt structured output space for semantic segmentation,''
  \emph{2018 IEEE/CVF Conference on Computer Vision and Pattern Recognition},
  pp. 7472--7481, 2018.

\bibitem{Liu2021PanoSfMLearnerSM}
M.~Liu, S.~Wang, Y.~Guo, Y.~He, and H.~Xue, ``Pano-sfmlearner: Self-supervised
  multi-task learning of depth and semantics in panoramic videos,'' \emph{IEEE
  Signal Processing Letters}, vol.~28, pp. 832--836, 2021.

\bibitem{Zhang2021DeepPanoContextP3}
C.~Zhang, Z.~Cui, C.~Chen, S.~Liu, B.~Zeng, H.~Bao, and Y.~Zhang,
  ``Deeppanocontext: Panoramic 3d scene understanding with holistic scene
  context graph and relation-based optimization,'' \emph{2021 IEEE/CVF
  International Conference on Computer Vision (ICCV)}, pp. 12\,612--12\,621,
  2021.

\bibitem{Wang2021DomainAS}
Q.~Wang, D.~Dai, L.~Hoyer, O.~Fink, and L.~V. Gool, ``Domain adaptive semantic
  segmentation with self-supervised depth estimation,'' \emph{2021 IEEE/CVF
  International Conference on Computer Vision (ICCV)}, pp. 8495--8505, 2021.

\bibitem{Zhang2017CurriculumDA}
Y.~Zhang, P.~David, and B.~Gong, ``Curriculum domain adaptation for semantic
  segmentation of urban scenes,'' \emph{2017 IEEE International Conference on
  Computer Vision (ICCV)}, pp. 2039--2049, 2017.

\bibitem{Li2019BidirectionalLF}
Y.~Li, L.~Yuan, and N.~Vasconcelos, ``Bidirectional learning for domain
  adaptation of semantic segmentation,'' \emph{2019 IEEE/CVF Conference on
  Computer Vision and Pattern Recognition (CVPR)}, pp. 6929--6938, 2019.

\bibitem{Murez2018ImageTI}
Z.~Murez, S.~Kolouri, D.~J. Kriegman, R.~Ramamoorthi, and K.~Kim, ``Image to
  image translation for domain adaptation,'' \emph{2018 IEEE/CVF Conference on
  Computer Vision and Pattern Recognition}, pp. 4500--4509, 2018.

\bibitem{Chen2019ProgressiveFA}
C.~Chen, W.~Xie, T.~Xu, W.~Huang, Y.~Rong, X.~Ding, Y.~Huang, and J.~Huang,
  ``Progressive feature alignment for unsupervised domain adaptation,''
  \emph{2019 IEEE/CVF Conference on Computer Vision and Pattern Recognition
  (CVPR)}, pp. 627--636, 2019.

\bibitem{Hoffman2016FCNsIT}
J.~Hoffman, D.~Wang, F.~Yu, and T.~Darrell, ``Fcns in the wild: Pixel-level
  adversarial and constraint-based adaptation,'' \emph{ArXiv}, vol.
  abs/1612.02649, 2016.

\bibitem{Luo2019TakingAC}
Y.~Luo, L.~Zheng, T.~Guan, J.~Yu, and Y.~Yang, ``Taking a closer look at domain
  shift: Category-level adversaries for semantics consistent domain
  adaptation,'' \emph{2019 IEEE/CVF Conference on Computer Vision and Pattern
  Recognition (CVPR)}, pp. 2502--2511, 2019.

\bibitem{MelasKyriazi2021PixMatchUD}
L.~Melas-Kyriazi and A.~K. Manrai, ``Pixmatch: Unsupervised domain adaptation
  via pixelwise consistency training,'' \emph{2021 IEEE/CVF Conference on
  Computer Vision and Pattern Recognition (CVPR)}, pp. 12\,430--12\,440, 2021.

\bibitem{eder2020tangent}
M.~Eder, M.~Shvets, J.~Lim, and J.-M. Frahm, ``Tangent images for mitigating
  spherical distortion,'' in \emph{Proceedings of the IEEE/CVF Conference on
  Computer Vision and Pattern Recognition}, 2020, pp. 12\,426--12\,434.

\bibitem{li2022omnifusion}
Y.~Li, Y.~Guo, Z.~Yan, X.~Huang, Y.~Duan, and L.~Ren, ``Omnifusion: 360
  monocular depth estimation via geometry-aware fusion,'' in \emph{Proceedings
  of the IEEE/CVF Conference on Computer Vision and Pattern Recognition}, 2022,
  pp. 2801--2810.

\bibitem{yang2021capturing}
K.~Yang, J.~Zhang, S.~Rei{\ss}, X.~Hu, and R.~Stiefelhagen, ``Capturing
  omni-range context for omnidirectional segmentation,'' in \emph{Proceedings
  of the IEEE/CVF Conference on Computer Vision and Pattern Recognition}, 2021,
  pp. 1376--1386.

\bibitem{wang2020differential}
Z.~Wang, M.~Yu, Y.~Wei, R.~Feris, J.~Xiong, W.-m. Hwu, T.~S. Huang, and H.~Shi,
  ``Differential treatment for stuff and things: A simple unsupervised domain
  adaptation method for semantic segmentation,'' in \emph{Proceedings of the
  IEEE/CVF Conference on Computer Vision and Pattern Recognition}, 2020, pp.
  12\,635--12\,644.

\bibitem{wang2021pyramid}
W.~Wang, E.~Xie, X.~Li, D.-P. Fan, K.~Song, D.~Liang, T.~Lu, P.~Luo, and
  L.~Shao, ``Pyramid vision transformer: A versatile backbone for dense
  prediction without convolutions,'' in \emph{Proceedings of the IEEE/CVF
  international conference on computer vision}, 2021, pp. 568--578.

\bibitem{armeni2017joint}
I.~Armeni, S.~Sax, A.~R. Zamir, and S.~Savarese, ``Joint 2d-3d-semantic data
  for indoor scene understanding,'' \emph{arXiv preprint arXiv:1702.01105},
  2017.

\bibitem{Cityscapes}
M.~Cordts, M.~Omran, S.~Ramos, T.~Rehfeld, M.~Enzweiler, R.~Benenson,
  U.~Franke, S.~Roth, and B.~Schiele, ``The cityscapes dataset for semantic
  urban scene understanding,'' in \emph{Proc. of the IEEE Conference on
  Computer Vision and Pattern Recognition (CVPR)}, 2016.

\bibitem{densepass}
C.~Ma, J.~Zhang, K.~Yang, A.~Roitberg, and R.~Stiefelhagen, ``Densepass: Dense
  panoramic semantic segmentation via unsupervised domain adaptation with
  attention-augmented context exchange,'' in \emph{2021 IEEE International
  Intelligent Transportation Systems Conference (ITSC)}.\hskip 1em plus 0.5em
  minus 0.4em\relax IEEE, 2021, pp. 2766--2772.

\end{thebibliography}
}

\end{document}